\documentclass[10pt,twocolumn,letterpaper]{article}

\usepackage[pagenumbers]{wacv} 

\usepackage{graphicx}
\usepackage{amsmath}
\usepackage{amssymb}
\usepackage{booktabs}
\usepackage{tabularx}

\usepackage[normalem]{ulem}
\usepackage{nicematrix,enumitem}
\usepackage{graphbox}

\usepackage[pagebackref,breaklinks,colorlinks]{hyperref}

\usepackage[capitalize]{cleveref}
\crefname{section}{Sec.}{Secs.}
\Crefname{section}{Section}{Sections}
\Crefname{table}{Table}{Tables}
\crefname{table}{Tab.}{Tabs.}

\def\wacvPaperID{1252} 
\def\confName{WACV}
\def\confYear{2025}

\begin{document}

\title{Face Anonymization Made Simple}

\author{
  Han-Wei Kung$^1$
  \and
  Tuomas Varanka$^2$\\
  \and
  Sanjay Saha$^3$\\
  \and
  Terence Sim$^3$\\
  \and
  Nicu Sebe$^1$\\
  \and
  $^1$University of Trento
  \and
  $^2$University of Oulu
  \and
  $^3$National University of Singapore
}

\twocolumn[{
  \renewcommand\twocolumn[1][]{#1}%
  \maketitle

  \begin{center}
    \centering
    \captionsetup{type=figure}

    \resizebox{.7\textwidth}{!}{
      \begin{tabular}{ 
          *{4}{>{\centering\arraybackslash}m{\dimexpr.25\linewidth-2\tabcolsep-.25\arrayrulewidth}}
        }
        \hline
        \multicolumn{1}{c|}{Original} & \multicolumn{3}{c}{Identity Anonymized} \\ 
        \hline
        \multicolumn{1}{c|}{\includegraphics[width={\dimexpr.25\linewidth}]{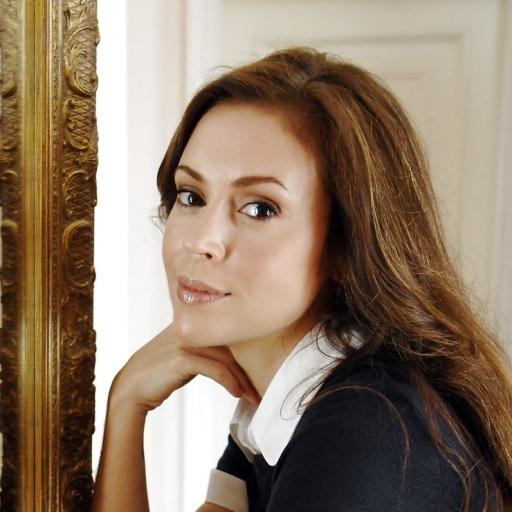}} & \multicolumn{1}{c}{\includegraphics[width={\dimexpr.25\linewidth}]{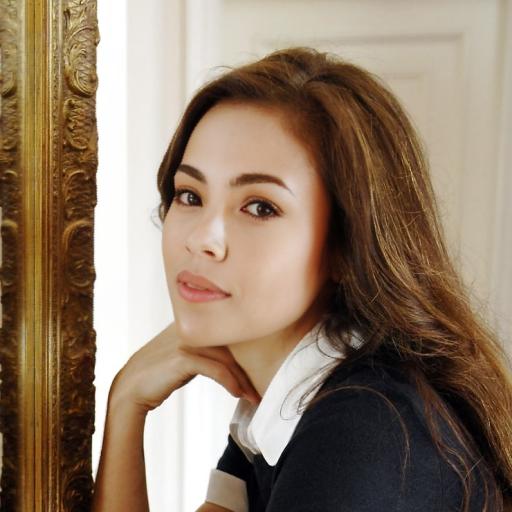}} & \multicolumn{1}{c}{\includegraphics[width={\dimexpr.25\linewidth}]{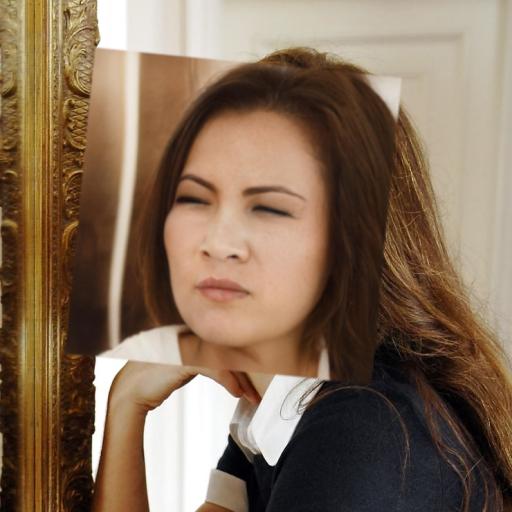}} & \multicolumn{1}{c}{\includegraphics[width={\dimexpr.25\linewidth}]{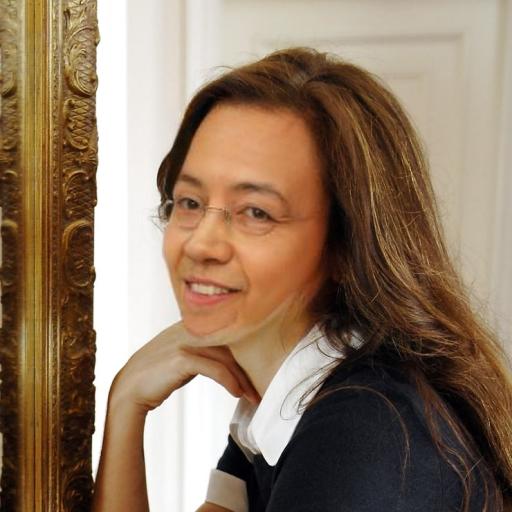}} \\
        \multicolumn{1}{c|}{} & Ours & FALCO~\cite{barattin2023attribute} & DP2~\cite{hukkelaas2023deepprivacy2} \\
        \hline
        \multicolumn{1}{c|}{\includegraphics[width={\dimexpr.25\linewidth}]{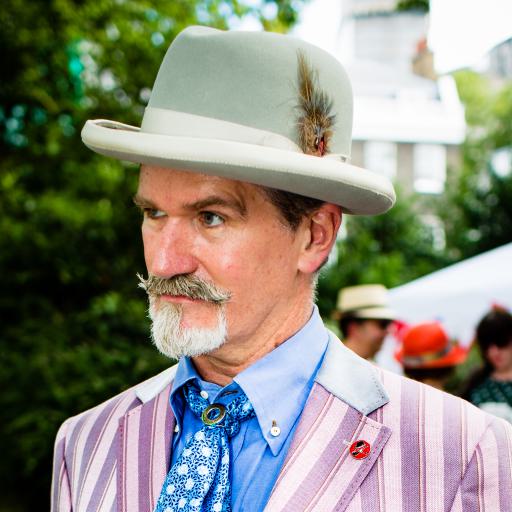}} & \multicolumn{1}{c}{\includegraphics[width={\dimexpr.25\linewidth}]{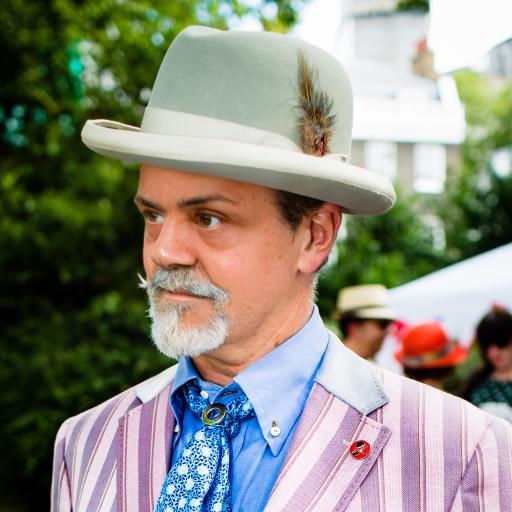}} & \multicolumn{1}{c}{\includegraphics[width={\dimexpr.25\linewidth}]{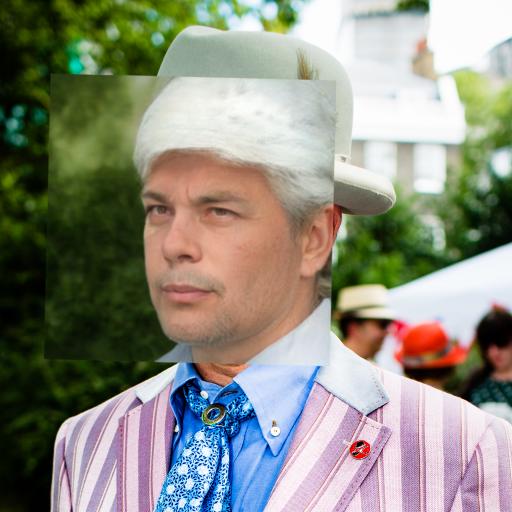}} & \multicolumn{1}{c}{\includegraphics[width={\dimexpr.25\linewidth}]{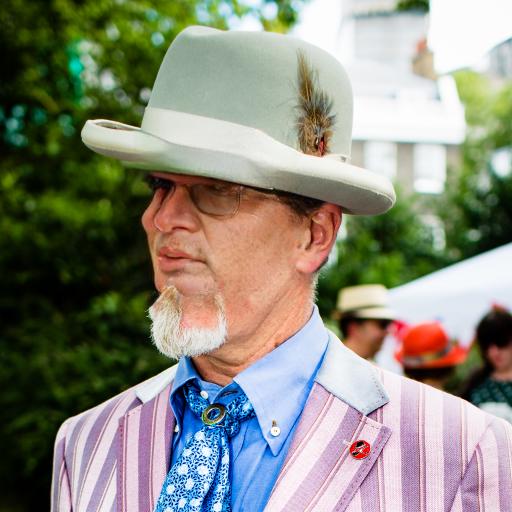}} \\
        \multicolumn{1}{c|}{} & Ours & RiDDLE~\cite{li2023riddle} & DP2~\cite{hukkelaas2023deepprivacy2} \\
        \hline
      \end{tabular}
    }

    \captionof{figure}{Our face anonymization technique preserves the original facial expressions, head positioning, eye direction, and background elements, effectively masking identity while retaining other crucial details. The anonymized face blends seamlessly into its original photograph, making it ideal for diverse real-world applications.}
  \end{center}
}]

\begin{abstract}
  Current face anonymization techniques often depend on identity loss calculated by face recognition models, which can be inaccurate and unreliable. Additionally, many methods require supplementary data such as facial landmarks and masks to guide the synthesis process. In contrast, our approach uses diffusion models with only a reconstruction loss, eliminating the need for facial landmarks or masks while still producing images with intricate, fine-grained details. We validated our results on two public benchmarks through both quantitative and qualitative evaluations. Our model achieves state-of-the-art performance in three key areas: identity anonymization, facial attribute preservation, and image quality. Beyond its primary function of anonymization, our model can also perform face swapping tasks by incorporating an additional facial image as input, demonstrating its versatility and potential for diverse applications. Our code and models are available at \url{https://github.com/hanweikung/face_anon_simple}.
\end{abstract}

\section{Introduction}
In the digital age, our identity and privacy are more vulnerable than ever. People have shared personal information and photos online over recent decades, while advancements in facial recognition technology have made it easier to identify individuals from a single image. This combination allows for the potential linking of our faces to personal information, posing a significant threat to our privacy and identity. In response, various regions have enacted privacy protection laws. These include the European Union's General Data Protection Regulation (GDPR)~\cite{gdpr}, California's Consumer Privacy Act, and Japan's amended Act on the Protection of Personal Information. Such legislation mandates that organizations implement security measures and maintain transparency in their handling of personal data.

Face anonymization is essential for protecting individuals in photos and videos, thereby reducing the risk of personal data being compromised or misused. Traditional methods like blurring and pixelation are common but have significant drawbacks. These techniques are vulnerable to reconstruction attacks~\cite{todt2022fant}, degrade image quality, and apply a uniform transformation across the image without considering which areas are most critical to anonymize.

These limitations make traditional methods impractical for professionals who need to preserve facial expressions and backgrounds. For example, medical practitioners may need to anonymize patient images for case studies or research while retaining crucial facial cues that indicate symptoms. In creative fields, documentary filmmakers might want to protect interviewees' privacy without losing the narrative impact of their facial expressions and reactions. They may also wish to replace an interviewee's face with a specific virtual identity to enhance storytelling clarity. In contrast, recent advances in deep learning have led to more effective anonymization techniques that enhance both privacy protection and usability. Generative Adversarial Networks (GANs)~\cite{goodfellow2020generative}, in particular, can anonymize faces by replacing the original with computer-generated alternatives~\cite{hukkelaas2019deepprivacy,maximov2020ciagan,ciftci2023my,sun2018natural}. However, these methods are not without challenges. Some fail to produce natural-looking faces~\cite{maximov2020ciagan}, while others~\cite{hukkelaas2019deepprivacy} struggle to preserve crucial elements like facial expressions, eye direction, head orientation, background details, clothing, and accessories. These limitations greatly restrict the practical application of these techniques.

This paper presents a diffusion-based method for face anonymization. Our goal is to ensure that de-identified facial images remain useful for facial analysis tasks, including pose estimation, eye-gaze tracking, and expression recognition, as well as for broader uses such as interviews and films. Therefore, we approach face anonymization similarly to face swapping, aiming to generate an image where a person's face is replaced by another person's face while maintaining the original facial expression, pose, eye gaze, and background. We designed a framework that initially performs realistic and seamless face swaps given both source and driving images. At its core is a denoising UNet architecture, similar to those used in text-to-image diffusion models, which generates the final output. We enhance this with an image feature extraction mechanism that transfers fine details from input images to the synthesized output throughout the diffusion process. The model is then trained in a dual setting: conditionally with a source image and unconditionally without a source image. This dual method allows the model to replace faces using one single image input. To create a distinct anonymized identity, the system reverses the original face's most distinctive features. This technique produces a believable anonymized face while preserving the original image's quality and essential facial characteristics.

In summary, our contributions are:
\begin{itemize}
  \item A convenient method that produces realistic anonymized faces while preserving attributes, without needing external data like facial landmarks or masks as required by existing techniques.
  \item A diffusion-based network that achieves good performance with a single, simple loss function, in contrast to GAN-based models requiring multiple, carefully designed loss functions.
  \item Simple control of the anonymization level using a single parameter.
  \item Versatility beyond anonymization, including the ability to perform face swapping tasks with an additional facial image input.
\end{itemize}

\section{Related Work}

\paragraph{Face Anonymization.}
Most deep learning-based image anonymization methods have been developed using GANs and target not only faces~\cite{zhai2022a3gan,ciftci2023my,rosberg2023fiva,barattin2023attribute,li2023riddle,hukkelaas2019deepprivacy,sun2018natural,sun2018hybrid,gu2020password,wen2023divide,dall2022graph,helou2023vera} but also bodies~\cite{hukkelaas2023deepprivacy2,ciftci2024my} and other objects~\cite{shvai2023adaptive}. In this study, we focus on face anonymization.

Many GAN-based face anonymization methods use conditional GANs as their foundation. These techniques typically require supplementary data to create anonymized faces. For example, IDeudemon~\cite{wen2023divide} uses face parsing maps or masks to segment image components, while Sun \etal~\cite{sun2018natural}'s method employs facial landmarks to guide face inpainting. CIAGAN~\cite{maximov2020ciagan} relies on masks and facial landmarks, and DeepPrivacy~\cite{hukkelaas2019deepprivacy} utilizes bounding boxes and facial landmarks. These methods depend on additional information, which can be a limitation if the required data are missing or flawed. In contrast, our approach does not rely on such auxiliary data to anonymize faces.

Other techniques like RiDDLE~\cite{li2023riddle} and FALCO~\cite{barattin2023attribute} use GAN inversion. They map facial images to the latent space of a pre-trained StyleGAN2~\cite{karras2020analyzing}, leveraging its capabilities to produce high-quality images. However, these techniques may inadvertently alter important identity-irrelevant details such as facial expressions, background, body parts, and accessories. Our method treats face anonymization similarly to face swapping and incorporates image feature extraction networks to capture detailed input features. This allows us to generate anonymized faces that seamlessly integrate with the existing image while preserving the overall integrity of the image.

StyleFace~\cite{luo2022styleface} embeds identity vectors from a pre-trained face recognition network into the StyleGAN2~\cite{karras2020analyzing} model's latent space, sampling random vectors for anonymization. While this approach generates realistic faces, it risks revealing the original identity if the sampled vector is too close to the original identity. In contrast, our model offers an adjustable anonymization degree, allowing users to control the distance between the input and generated images for effective anonymization.

\begin{figure*}[t]
  \centering
  \includegraphics[width=0.9\linewidth]{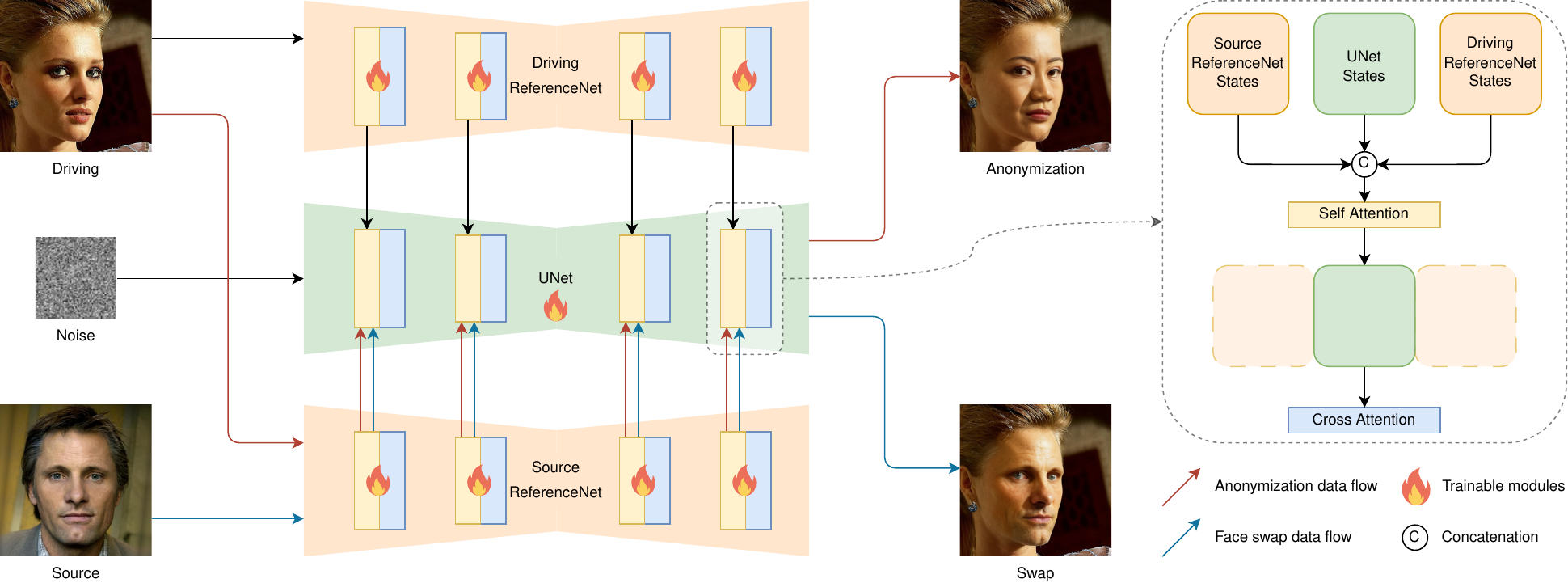}
  \caption{Our network leverages the face swapping mechanism for face anonymization. In both cases, the system encodes source and driving images into latent space and processes them through respective ReferenceNet models. These images are also encoded into intermediate embeddings that guide the UNet via cross-attention. The UNet incorporates states from both ReferenceNet models through concatenation, enabling the transfer of details from source and driving images through self-attention. Using these learned features and intermediate embeddings, the UNet generates the output image. For face anonymization, we use the same image as both source and driving input. However, we modify the intermediate embedding and state from the source ReferenceNet model to achieve the desired anonymization effect.}
  \label{fig:pipeline}
\end{figure*}

\paragraph{Face Swapping.}
Face swapping techniques can be categorized into two main approaches: source-oriented and target-oriented methods.

Source-oriented methods~\cite{yuan2023reliableswap,nirkin2019fsgan,li2023e4s} begin by transforming the source face to match the expression, pose, and lighting of the target face, and then replace the target image with this modified source face. For example, FSGAN~\cite{nirkin2019fsgan} employs a two-stage process: it first uses a reenactment network for expression and pose transfer, and then an inpainting network to blend the source face into the target image. Similarly, E4S~\cite{li2023e4s} uses face reenactment to align the source image with the target's pose, followed by swapping faces using masks and texture information. However, these methods are sensitive to the source image; exaggerated expressions or extreme poses in the source can adversely affect the swapping result.

Target-oriented methods~\cite{Deepfakes_FaceSwap,Iperov_DeepFaceLab,shiohara2023blendface,rosberg2023facedancer,chen2020simswap,gao2021information,li2019faceshifter,wang2021hififace,ren2023reinforced}, on the other hand, modify the features of the target image to incorporate the source identity. Some of these methods~\cite{Deepfakes_FaceSwap,Iperov_DeepFaceLab}, based on autoencoder architecture, can swap between specific identities, while others, like GAN-based approaches~\cite{chen2020simswap,shiohara2023blendface,rosberg2023facedancer,gao2021information,li2019faceshifter,wang2021hififace,ren2023reinforced}, can generalize to various identities by merging the source identity and target attributes at the feature level. For example, SimSwap~\cite{chen2020simswap} offers an efficient framework for high-fidelity face swapping by injecting the source identity into the target features and using a weak feature matching loss to maintain attributes. These methods are more adept at handling variations in the source face compared to source-oriented methods. However, they often struggle to balance competing objectives, such as reconstruction loss and identity loss.

Our diffusion-based approach differs from these methods by relying on a single reconstruction loss for simplicity, while still generating images that both look natural in the target context and preserve the source face's identity.

\section{Methodology}
Our approach to face anonymization is similar to face swapping, but with a key difference. In face swapping, two images are used: a driving image (containing the face to be replaced) and a source image (providing the new face). Our face anonymization method, however, requires only one input image. Therefore, we developed a framework that initially learns to perform realistic face swaps using both driving and source images. We then expanded this model to work in two scenarios simultaneously: one where a source image is provided, and the other where no source image is available. This dual training allows the model to generate a new face even when given just one image. The result is a system that can synthesize a convincing, anonymous face while preserving the original image's facial expression, head posture, gaze direction, and surrounding elements. This achieves our main goal: replacing a person's face without revealing their identity or compromising the image's overall quality.

\subsection{A New Paradigm}

We aim to address several common limitations in current face anonymization and face swapping techniques.

First, while facial landmarks and masks provide a structured approach for face anonymization~\cite{maximov2020ciagan,hukkelaas2019deepprivacy,sun2018natural,wen2023divide} and face swapping~\cite{wang2021hififace,zhu2024stableswap,zhao2023diffswap,li2023e4s,nirkin2019fsgan}, they have inherent limitations that can compromise the quality, realism, and flexibility of generated images. These methods identify major features like the eyes, nose, and mouth but miss finer details such as skin texture and nuanced expressions. This oversimplification results in less realistic and detailed facial representations compared to methods that consider pixel-level information. Additionally, the quality of the generated face heavily relies on the accurate detection of landmarks and masks; inaccuracies can lead to distorted or unrealistic faces. Moreover, facial landmarks and masks struggle to effectively capture dynamic expressions and poses, limiting the ability to generate faces with a wide range of emotions and orientations.

\begin{figure}
  \footnotesize
  \begin{tabular}{*{5}{>{\centering\arraybackslash}m{\dimexpr.2\linewidth-2\tabcolsep}}}
    \hline
    Original & \multicolumn{3}{|c|}{ Different IDs } & Same ID \\ [\defaultaddspace]
    \multicolumn{5}{@{}c@{}}{\includegraphics[width=\linewidth]{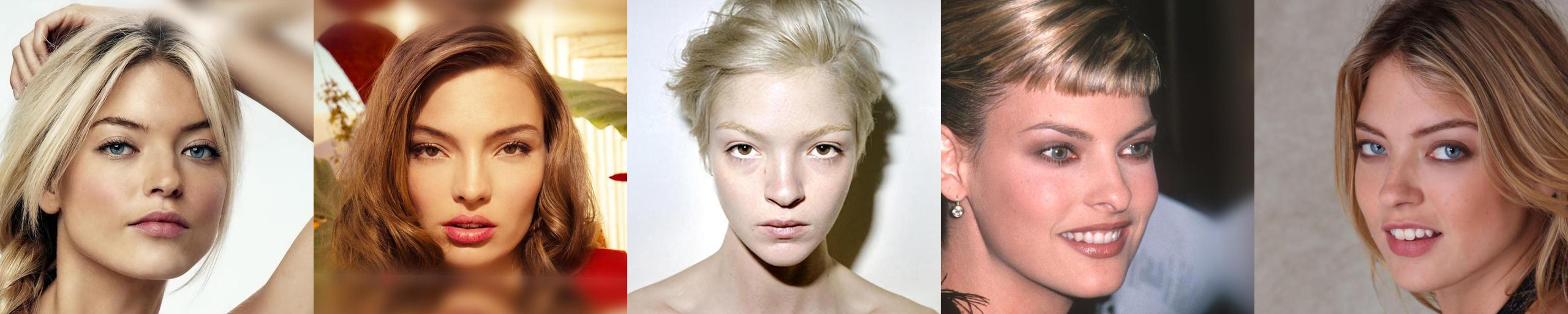}} \\ 
    \multicolumn{1}{@{}c@{}}{ID Dist.} & 0.406 & 0.448 & 0.462 & 0.637 \\ [\defaultaddspace]
    \multicolumn{5}{@{}c@{}}{\includegraphics[width=\linewidth]{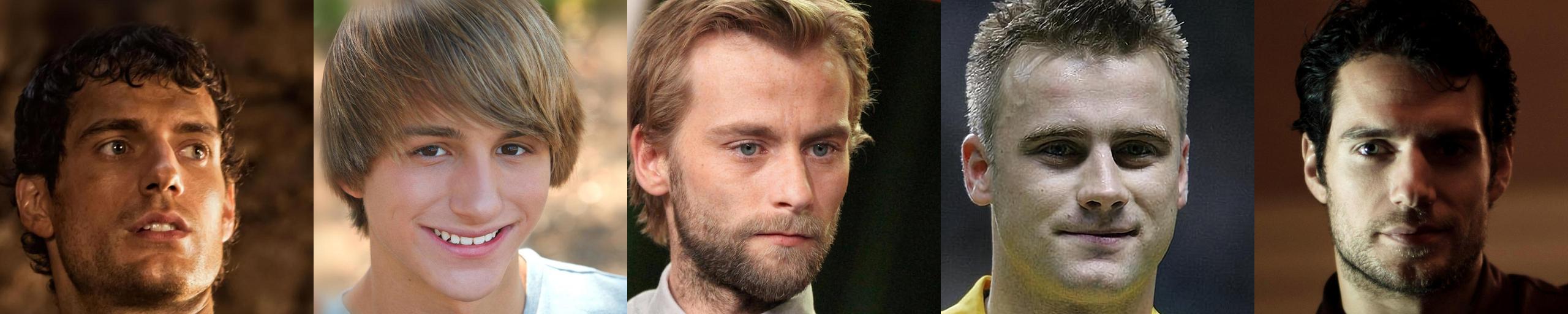}} \\
    \multicolumn{1}{@{}c@{}}{ID Dist.} & 0.577 & 0.577 & 0.600 & 0.636 \\ [\defaultaddspace]
    \multicolumn{1}{@{}c@{}}{} & \multicolumn{4}{@{}c@{}}{\includegraphics[width=.8\linewidth]{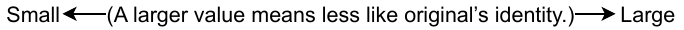}} \\
  \end{tabular}
  \caption{For each row, we show the identity distance of each image from the original image in that row, as calculated by the ArcFace~\cite{deng2019arcface} recognition model. The results indicate that the recognition model can generate inaccurate identity distances. It may assign a greater identity distance to two images of the same person than to two images of different people due to variations in head pose, facial expressions, or lighting conditions.}
  \label{fig:arcface}
\end{figure}

Second, using ArcFace~\cite{deng2019arcface}, a loss function in deep face recognition models, to learn discriminative facial features for face anonymization~\cite{barattin2023attribute,li2023riddle,rosberg2023fiva,zhai2022a3gan} can have drawbacks. The biases in these encoded features can negatively affect the quality of the anonymized faces. As shown in \cref{fig:arcface}, ArcFace~\cite{deng2019arcface} can sometimes produce misleading identity distances, indicating greater distance between two images of the same person than between two images of different individuals. These errors typically stem from variations in pose, lighting, facial expressions, occlusions, or image quality.

Lastly, training models for face swapping often involves optimizing multiple loss functions, such as reconstruction loss and identity loss, to address different aspects of the output. However, these losses can sometimes conflict, leading to suboptimal results. This issue often arises from insufficient disentanglement between identity and non-identity features. Methods that prioritize preserving the source identity, like those using 3D priors~\cite{wang2021hififace}, often lose the target's non-identity details. Conversely, approaches like Faceshifter~\cite{li2019faceshifter} and DiffSwap~\cite{zhao2023diffswap}, which focus on preserving the target's low-level attributes, risk allowing the target's facial identity to appear in the final swapped image.

To overcome these limitations, we use networks that capture and utilize pixel-level information, enhancing the quality of the generated faces without relying on additional facial landmarks or masks. Previous research~\cite{hu2024animate,wang2024vividpose,alzayer2024magic,xue2024follow} has shown that these networks effectively preserve the fine-grained details of input images. Additionally, we simplify the training process of our networks by employing a single mean squared error loss function, avoiding the complexities associated with multiple loss functions and the dependence on facial features encoded by face recognition models. This approach offers several advantages, including simplicity, stability, and improved quality.

\subsection{Architecture}

As illustrated in \cref{fig:pipeline}, our architecture uses the Latent Diffusion Model~\cite{rombach2022high}, based on a UNet structure, to produce the final output images. Stacked on top of this UNet are two ReferenceNet~\cite{hu2024animate} models that transfer fine-grained details from the input images. The first ReferenceNet model, which we call the source ReferenceNet model, takes the source images as input. These images provide information about the desired identity to be transferred. The second model, named the driving ReferenceNet model, takes the driving images as input. These images set the non-identity related conditions, such as pose, expression, and background details.

ReferenceNet shares the same structure as UNet. It captures details from input images and modulates the UNet through self-attention at each diffusion step. The process unfolds as follows: First, an input image is encoded into latent space using the CLIP~\cite{radford2021learning} image encoder and then passed to ReferenceNet. Within each attention module of ReferenceNet, self-attention is applied to extract features from the CLIP-encoded image. These extracted features serve as input states for corresponding attention modules in the UNet. Specifically, the self-attention modules within the UNet receive the concatenated states from all three models---the two ReferenceNet models and the UNet itself. The output from these modules is split into three parts, with one part sent to the UNet's cross-attention module for further processing. This step is also depicted in \cref{fig:pipeline}.

Our architectural design offers three main advantages. First, due to their similar network structures, ReferenceNet can retain the extensive knowledge and capabilities that the UNet acquired from a large dataset by adopting its pre-trained weights. This approach prevents the training of ReferenceNet from compromising the UNet's performance and enhances both ReferenceNet's performance and training efficiency. Second, the UNet can utilize knowledge learned by ReferenceNet because of their structural similarities and shared initialization weights. This allows the UNet to extract and incorporate relevant features from ReferenceNet during training, as both networks operate in a shared feature space. Finally, by separating the data flows for source and driving images, the UNet can more effectively identify which features of the driving image to retain and which to replace with those from the source image. This clear distinction is crucial for synthesizing the final output image accurately.

\subsection{Anonymization}

Our framework's training method enables the UNet to selectively learn identity information from the source ReferenceNet model and non-identity-related information from the driving ReferenceNet model. The UNet then combines these two types of information to synthesize a new facial image. To anonymize a facial image, we use the same image as input for both source and driving ReferenceNet models, while adjusting intermediate inputs to the source ReferenceNet and UNet models. Specifically, we modify two key components:

\begin{enumerate}
  \item Intermediate image embedding. We adjust the intermediate image embedding from the image encoder using this equation:

    \begin{equation}
      Z_{img}' = (1 - d) \cdot Z_{img}
      \label{eq:embedding}
    \end{equation}

    Here, $Z_{img}'$ is the adjusted embedding, $d$ controls the degree of anonymization, and $Z_{img}$ is the original embedding. As $d$ increases, more identity information is removed from $Z_{img}'$. This adjusted embedding influences both the source ReferenceNet (via self-attention) and the UNet (via cross-attention).

  \item Source ReferenceNet state. We modify the state of the source ReferenceNet using this equation:

    \begin{equation}
      S' = (1 - d) \cdot S_{cond} + d \cdot S_{uncond}
      \label{eq:state}
    \end{equation}

    $S'$ is the modified state, $d$ is the same factor controlling the degree of anonymization, $S_{cond}$ is the conditional state (with identity information), and $S_{uncond}$ is the unconditional state (without identity information). As $d$ increases, $S'$ shifts further from the conditional state towards the unconditional state. The modified state $S'$ is then incorporated into the UNet's intermediate layers using self-attention.

\end{enumerate}

Simply put, the equations demonstrate that by increasing the parameter $d$, the original identity is gradually removed from the resulting image while an unknown identity is progressively introduced. This process transforms the original identity into a different one, effectively achieving the desired anonymization.

\section{Experiments}

This section includes our experimental setup, procedures, findings, and approaches used to analyze our results.

\subsection{Implementation Details}

We trained our model using three datasets: CelebRef-HQ~\cite{li2022learning}, CelebA-HQ~\cite{karras2017progressive}, and FFHQ~\cite{karras2019style}. Face recognition~\cite{schroff2015facenet} was used to identify images of the same person, and for each identity, two images were randomly selected: one as the source and one as the ground truth. A synthesized driving image was then generated by using a state-of-the-art face-swapping model~\cite{Guo_InsightFace_Swapper} to replace the face in the ground truth image with another person's face. These three images---the source, synthesized driving, and ground truth---were used to train our model to learn identity changes. For a detailed breakdown of the number of images used in training, please refer to our supplementary material.

The ReferenceNet models and the UNet were initialized from a pre-trained Stable Diffusion~\cite{rombach2022high} v2.1 model. To incorporate classifier-free guidance~\cite{ho2022classifier}, we applied the unconditional mode to a random 10\% of the training data, while the conditional mode was used for the remaining 90\%.

During training, we discovered that focusing solely on the attention modules in the ReferenceNet model was as effective as training the entire model. This finding aligns with our understanding that these attention layers play a crucial role in shaping the structure and content of the generated images. As a result, we chose to optimize only the weights of the UNet and the attention modules in the ReferenceNet models. This targeted strategy allowed us to streamline our training process while maintaining effectiveness. We trained the model at a final output resolution of 512 $\times$ 512 over 435,000 steps. The training utilized the AdamW~\cite{loshchilov2017decoupled} optimizer with a batch size of 1 and 8 accumulation steps, maintaining a fixed learning rate of 1e-5. This process was conducted on two A6000 GPUs.

We also observed that using only synthesized images as driving images led to a problem where our model performed well only when the driving image was synthesized. To enhance performance and generalization, we adopted strategies from curriculum learning~\cite{bengio2009curriculum}. Initially, we trained the model with both real and synthesized driving images. When the driving image was real, we used its face-swapped counterpart as the ground truth and an image of the person originally used to swap the face in the driving image as the source. As training progressed, we transitioned to using only synthesized images as driving images and fine-tuned the model solely on real images as ground truth. This approach allows the model to first learn fundamental representations from a diverse set of data and then improve its capability to generate more realistic images.

Throughout this study, we maintained consistent parameters for image generation. We used the DDPM~\cite{ho2020denoising} algorithm with 200 denoising steps and a guidance scale value~\cite{ho2022classifier} of 4.0 for all examples presented in this paper.

\subsection{Achieving Diverse Anonymization Results}

\begin{figure}[t]
  \footnotesize
    \begin{tabular}{*{5}{>{\centering\arraybackslash}m{\dimexpr.2\linewidth-2\tabcolsep}}}
      \hline
      Original & \(d=0.3\) & \(d=0.6\) & \(d=0.9\) & \(d=1.2\) \\ [\defaultaddspace]
      \multicolumn{5}{@{}c@{}}{\includegraphics[width=\linewidth]{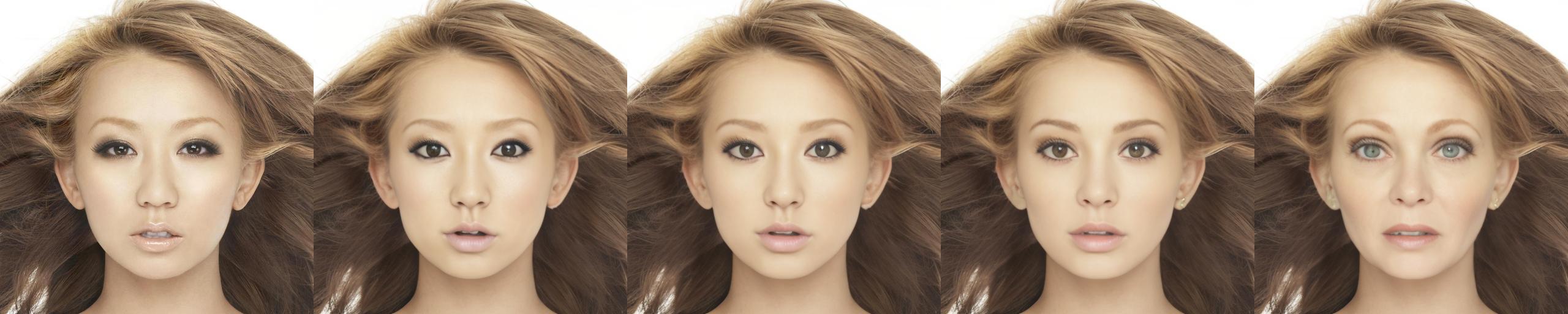}} \\
      \multicolumn{1}{@{}c@{}}{ID Dist.} & 0.151 & 0.262 & 0.782 & 1.080 \\ [\defaultaddspace]
      \multicolumn{5}{@{}c@{}}{\includegraphics[width=\linewidth]{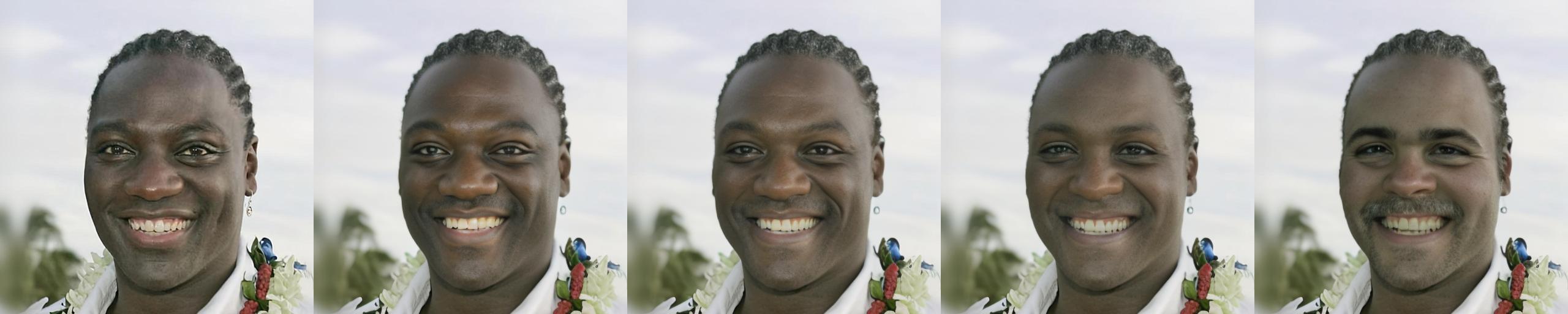}} \\
      \multicolumn{1}{@{}c@{}}{ID Dist.} & 0.208 & 0.281 & 0.408 & 0.951 \\ [\defaultaddspace]
      \multicolumn{1}{@{}c@{}}{} & \multicolumn{4}{@{}c@{}}{\includegraphics[width=.8\linewidth]{./images/arcface/arrow.pdf}} \\
    \end{tabular}
    \caption{Facial images generated with different degrees of anonymization. Each generated image reflects a different degree of anonymity applied to the original face. Alongside each generated image is a cosine distance score, calculated using the FaceNet~\cite{schroff2015facenet} recognition model. This score quantifies how different the anonymized face is from the original in terms of identity features.}
  \label{fig:anon_deg}
\end{figure}

\begin{figure}[t]
  \footnotesize
    \begin{tabular}{*{5}{>{\centering\arraybackslash}m{\dimexpr.2\linewidth-2\tabcolsep}}}
      \hline
      Original & Seed 32 & Seed 56 & Seed 68 & Seed 81 \\ [\defaultaddspace]
      \multicolumn{5}{@{}c@{}}{\includegraphics[width=\linewidth]{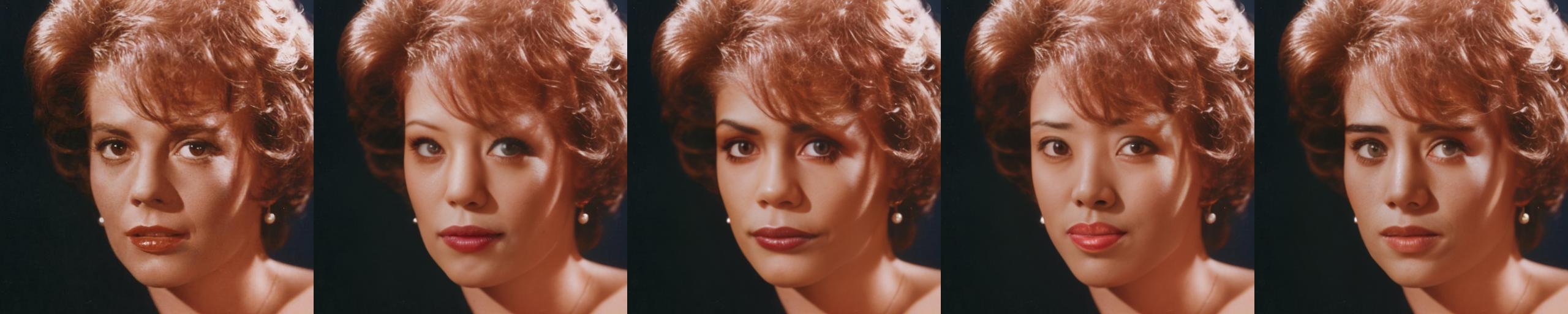}} \\
      \multicolumn{1}{@{}c@{}}{ID Dist.} & 0.578 & 0.444 & 0.986 & 0.568 \\ [\defaultaddspace]
      \multicolumn{5}{@{}c@{}}{\includegraphics[width=\linewidth]{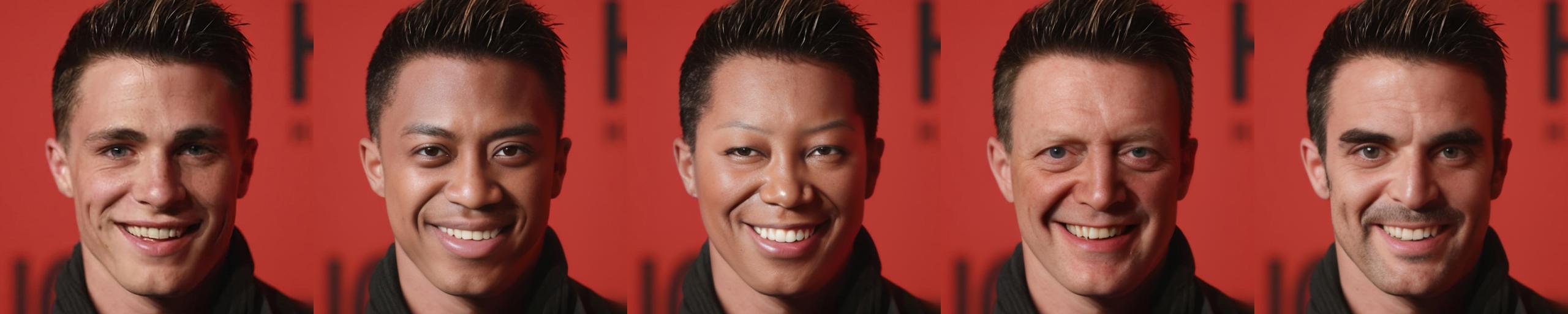}} \\
      \multicolumn{1}{@{}c@{}}{ID Dist.} & 0.860 & 0.888 & 0.865 & 0.906 \\
    \end{tabular}
    \caption{Various anonymized versions created from a single original identity, each using a different integer seed value. For each anonymized version, we present the cosine distance from its original identity, calculated using the FaceNet~\cite{schroff2015facenet} recognition model.}
  \label{fig:anon_var}
\end{figure}

Two methods allow us to vary anonymization results. First, we can modify the floating-point value $d$, defined in \cref{eq:embedding,eq:state}, which controls the anonymization intensity. Higher $d$ values produce images that deviate more from the original, as \cref{fig:anon_deg} demonstrates. When $d$ surpasses 1, the process moves in the opposite direction of the original identity's defining characteristics, ensuring the anonymized identity is not overly similar to the original. Second, we can use different integer seed values. This change introduces different initial Gaussian noise, leading to varied outcomes, as shown in \cref{fig:anon_var}.

\subsection{Baseline Comparisons}

\paragraph{Baselines.}
We benchmarked our model against three leading face anonymization methods (DP2~\cite{hukkelaas2023deepprivacy2}, FALCO~\cite{barattin2023attribute}, and RiDDLE~\cite{li2023riddle}) and three leading face swapping methods (DiffSwap~\cite{zhao2023diffswap}, BlendFace~\cite{shiohara2023blendface}, and InSwapper~\cite{Guo_InsightFace_Swapper}). For evaluation, we used images not included in the training datasets. Specifically, we selected 1,000 images each from CelebA-HQ~\cite{karras2017progressive} and FFHQ~\cite{karras2019style}, totaling 2,000 images for testing.

\paragraph{Evaluation Metrics.}
We evaluate the generated facial images using several metrics: re-identification rate, face shape distance, pose distance, gaze distance, expression distance, and image quality.

To calculate the re-identification rate, we extract identity vectors using the FaceNet~\cite{schroff2015facenet} recognition model and compute the cosine similarity to measure identity distance. For each generated face in the test set, we find the most similar face within the same test set. If this face matches the original face used for generation, we increment the re-identification count by one; otherwise, the count remains unchanged.

Face shape and expression distances are assessed using a face reconstruction model~\cite{deng2019accurate}. This model predicts 3DMM~\cite{blanz2023morphable} coefficients for both generated and original faces, allowing us to calculate the L2 Euclidean Distance between these coefficients.

For pose distance, we use a head pose estimation model~\cite{ruiz2018fine} to predict the orientation of both the generated and original faces. We then calculate the quaternion angular distance between these orientations. Gaze distance is computed similarly. We employ a gaze estimation model~\cite{abdelrahman2023l2cs} to predict the gaze direction of both the generated and original faces, then calculate the quaternion angular distance between these predicted directions.

Image quality is measured using an Image Quality Assessment (IQA) network~\cite{chen2024topiq} specifically trained on a face IQA dataset~\cite{su2023going}, which is ideal for evaluating the quality of facial images.

\paragraph{Quantitative Comparison.}

\begin{table*}[ht]
  \resizebox{\textwidth}{!}{
    \begin{tabular}{l|cc|cc|cc|cc|cc|cc}
      & \multicolumn{4}{c|}{Identity Distance} & \multicolumn{6}{c|}{Attribute Distance} & \multicolumn{2}{c}{Image Quality} \\
      \cline{2-13}
      & \multicolumn{2}{c|}{Re-ID $\downarrow$} & \multicolumn{2}{c|}{Shape $\uparrow$} & \multicolumn{2}{c|}{Pose $\downarrow$} & \multicolumn{2}{c|}{Gaze $\downarrow$} & \multicolumn{2}{c|}{Expression $\downarrow$} & \multicolumn{2}{c}{Face IQA $\uparrow$} \\
      & CelebA-HQ & FFHQ & CelebA-HQ & FFHQ & CelebA-HQ & FFHQ & CelebA-HQ & FFHQ & CelebA-HQ & FFHQ & CelebA-HQ & FFHQ \\
      \hline
      DP2~\cite{hukkelaas2023deepprivacy2} & 0.020 & 0.046 & 30.297 & 29.837 & 0.140 & 0.194 & 0.244 & 0.252 & 10.139 & \underline{9.613} & 0.459 & 0.480 \\
    FALCO~\cite{barattin2023attribute} & \textbf{0.005} & - & 31.816 & - & 0.088 & - & 0.258 & - & \underline{9.290} & - & \textbf{0.757} & - \\
      RiDDLE~\cite{li2023riddle} & - & \textbf{0.007} & - & \underline{36.624} & - & 0.090 & - & 0.220 & - & 10.018 & - & 0.571 \\
      Ours \((d=1.2)\) & 0.053 & 0.098 & \underline{33.046} & 28.971 & \textbf{0.048} & \textbf{0.047} & \textbf{0.161} & \textbf{0.166} & \textbf{8.256} & \textbf{7.769} & 0.701 & \underline{0.698} \\
      Ours \((d=1.4)\) & \underline{0.008} & \underline{0.039} & \textbf{53.244} & \textbf{41.695} & \underline{0.074} & \underline{0.061} & \underline{0.190} & \underline{0.206} & 13.125 & 10.899 & \underline{0.707} & \textbf{0.704} \\
    \end{tabular}
  }
  \caption{Quantitative results on the task of face anonymization for CelebA-HQ~\cite{karras2017progressive} and FFHQ~\cite{karras2019style} test sets, with the best results highlighted in bold and the second-best results underlined.}
  \label{tab:anon_quan}
\end{table*}

The quantitative results in \cref{tab:anon_quan} demonstrate our model's performance in face anonymization in comparison to baseline methods. We did not include the quantitative results for FALCO~\cite{barattin2023attribute} on the FFHQ~\cite{karras2019style} test set and RiDDLE~\cite{li2023riddle} on the CelebA-HQ~\cite{karras2017progressive} test set, as they require additional information that is not readily available. For quantitative results related to face swapping, please see our supplementary material.

\Cref{tab:anon_quan} indicates that our model, with $d=1.4$, excels in producing faces with highly distinct shapes while maintaining the original pose and gaze across both datasets. Conversely, when we set $d$ to a smaller value of 1.2, our model best preserves all three original facial attributes (pose, gaze, and expressions) across both datasets. However, this smaller $d$ comes with lower re-identification performance and face shapes more similar to the original. Generally, a smaller $d$ value improves attribute preservation, but results in lower re-identification performance and more similar face shapes. This is expected, as the generated image remains closer to the original.

We recognize that our method does not achieve the lowest re-identification rates compared to FALCO~\cite{barattin2023attribute} and RiDDLE~\cite{li2023riddle} when assessed by the FaceNet~\cite{schroff2015facenet} recognition model. We examined the cases where the recognition model successfully traced our model's outputs back to their original images. Many involved subjects from underrepresented groups in our training data, particularly infants and ethnic minorities like Asian individuals. This lack of representation led to poorer model performance in these scenarios. This data imbalance also explains why our model performs better on the CelebA-HQ~\cite{karras2017progressive} dataset compared to FFHQ~\cite{karras2019style}, as the former contains fewer examples of infants and minority groups. In comparison, RiDDLE~\cite{li2023riddle} achieves the lowest re-identification rate on the FFHQ~\cite{karras2019style} dataset, as it explicitly uses an identity loss term to distinguishes between real and anonymized faces. However, it also relies on several additional loss terms to preserve non-identity-relevant facial attributes and background. The use of multiple loss terms can lead to conflicts between different objectives, potentially resulting in less-than-ideal outcomes.

Regarding image quality, our model ranks second behind FALCO~\cite{barattin2023attribute}. This may be due to FALCO's~\cite{barattin2023attribute} ability to natively generate higher resolution images (1024 $\times$ 1024) compared to our model's native resolution of 512 $\times$ 512. While the SDXL~\cite{podell2023sdxl} model allows us to create images exceeding 512 $\times$ 512 resolution, training and testing such larger models require significantly more GPU memory, which is currently beyond our available resources.

\begin{figure}[ht]
  \scriptsize
  \centering
  \begin{tabular}{*{5}{>{\centering\arraybackslash}m{\dimexpr.2\linewidth-2\tabcolsep}}}
    \multicolumn{5}{@{}c@{}}{\includegraphics[width=.95\linewidth]{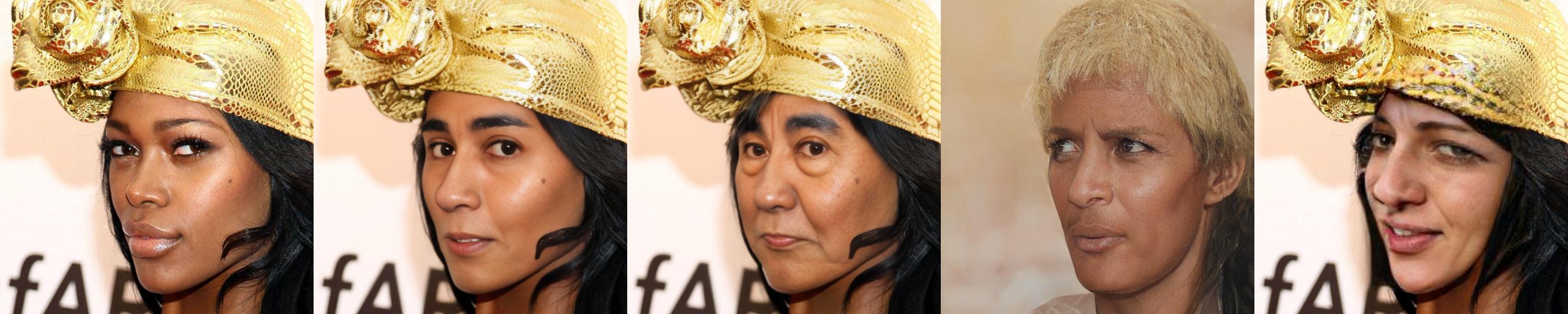}} \\
    \multicolumn{5}{@{}c@{}}{\includegraphics[width=.95\linewidth]{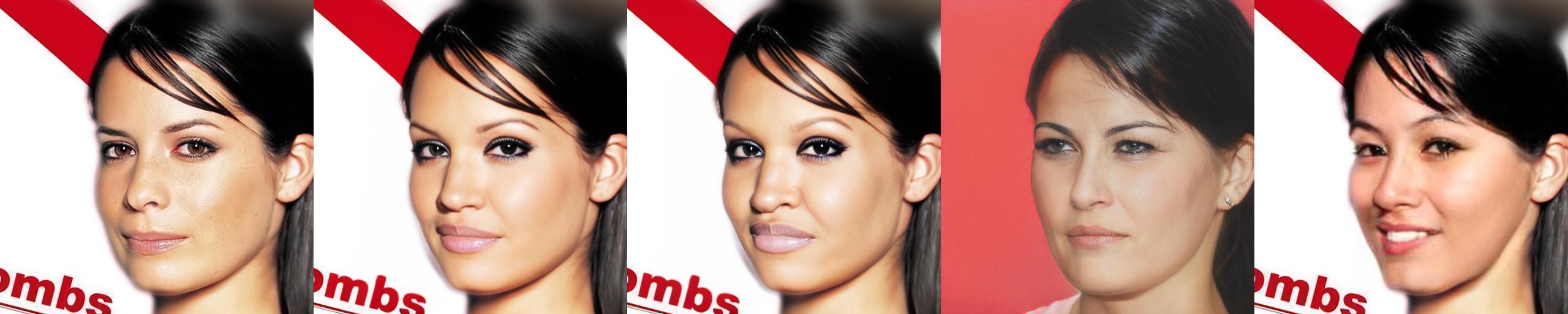}} \\
    \multicolumn{5}{@{}c@{}}{\includegraphics[width=.95\linewidth]{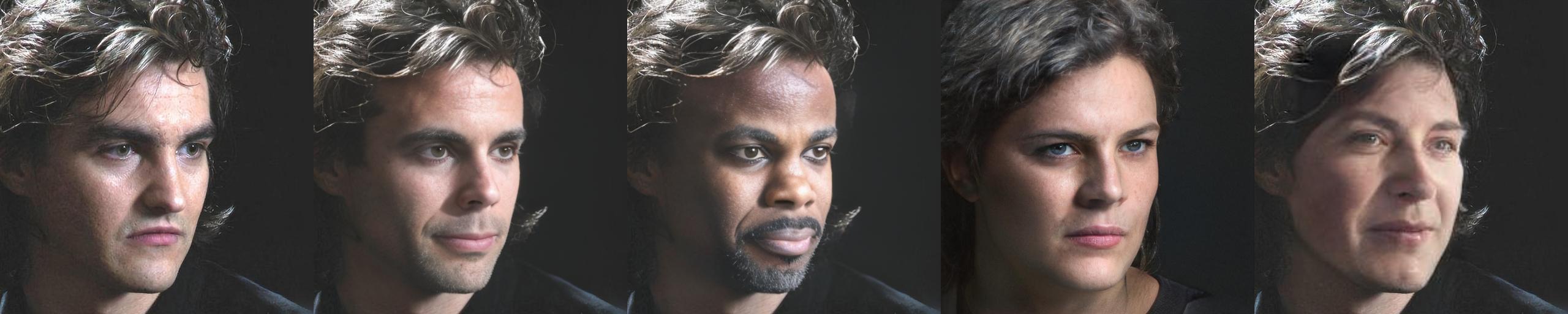}} \\
    \multicolumn{5}{@{}c@{}}{\includegraphics[width=.95\linewidth]{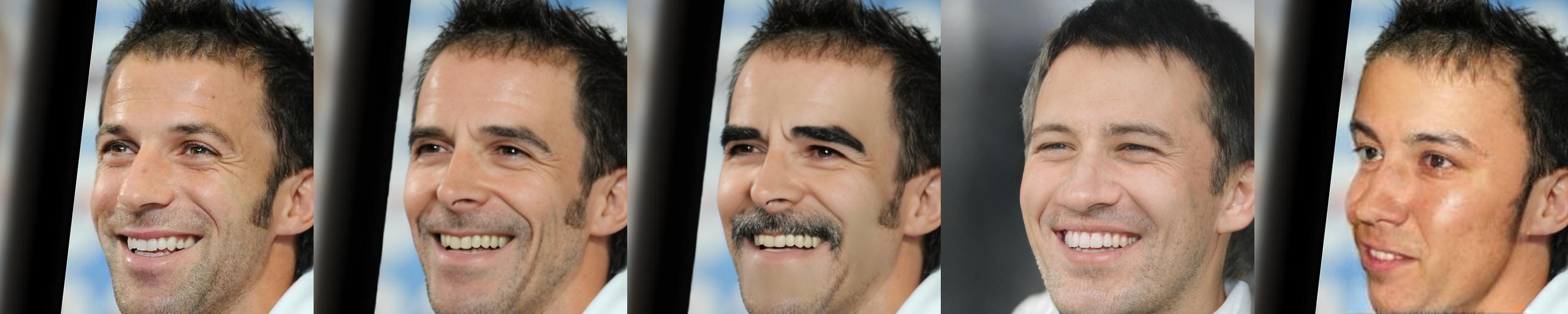}} \\
    Original & Ours \((d=1.2)\) & Ours \((d=1.4)\) & FALCO~\cite{barattin2023attribute} & DP2~\cite{hukkelaas2023deepprivacy2} \\
  \end{tabular}
  \caption{Qualitative results for the face anonymization task for the CelebA-HQ~\cite{karras2017progressive} test set.}
  \label{fig:anon_qual_cele}
\end{figure}

\begin{figure}[ht]
  \scriptsize
  \centering
  \begin{tabular}{*{5}{>{\centering\arraybackslash}m{\dimexpr.2\linewidth-2\tabcolsep}}}
    \multicolumn{5}{@{}c@{}}{\includegraphics[width=.95\linewidth]{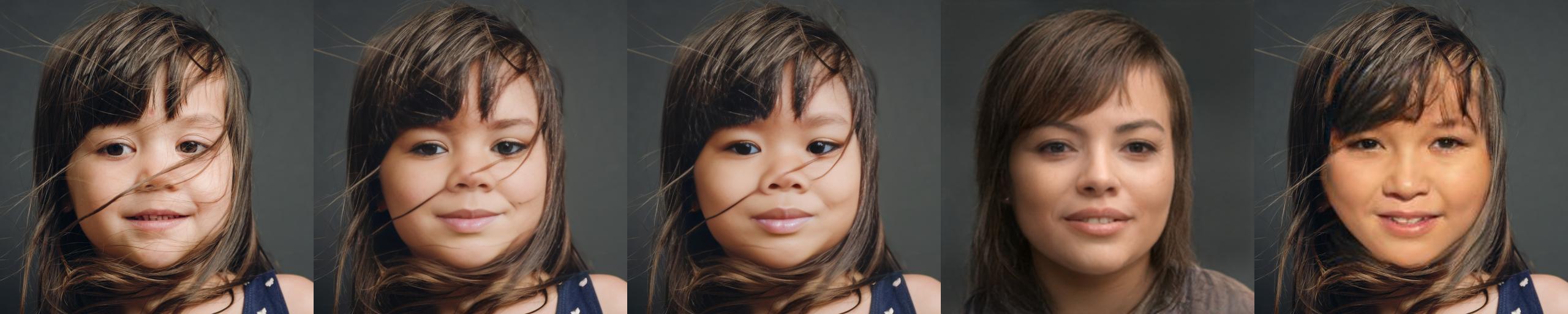}} \\
    \multicolumn{5}{@{}c@{}}{\includegraphics[width=.95\linewidth]{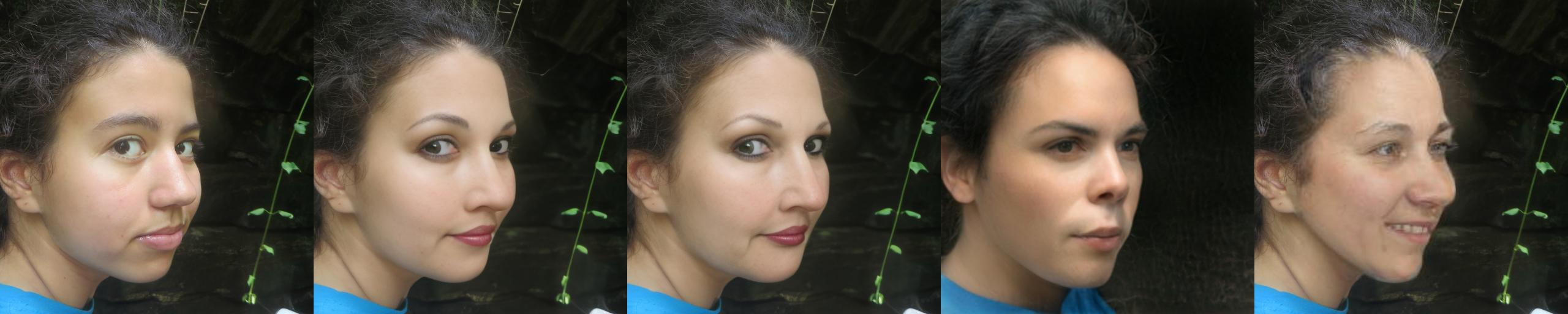}} \\
    \multicolumn{5}{@{}c@{}}{\includegraphics[width=.95\linewidth]{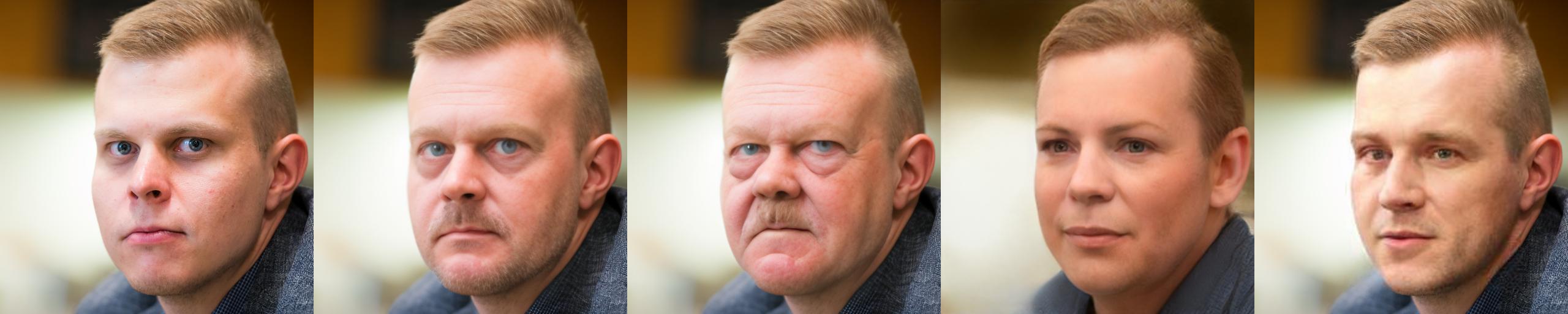}} \\
    \multicolumn{5}{@{}c@{}}{\includegraphics[width=.95\linewidth]{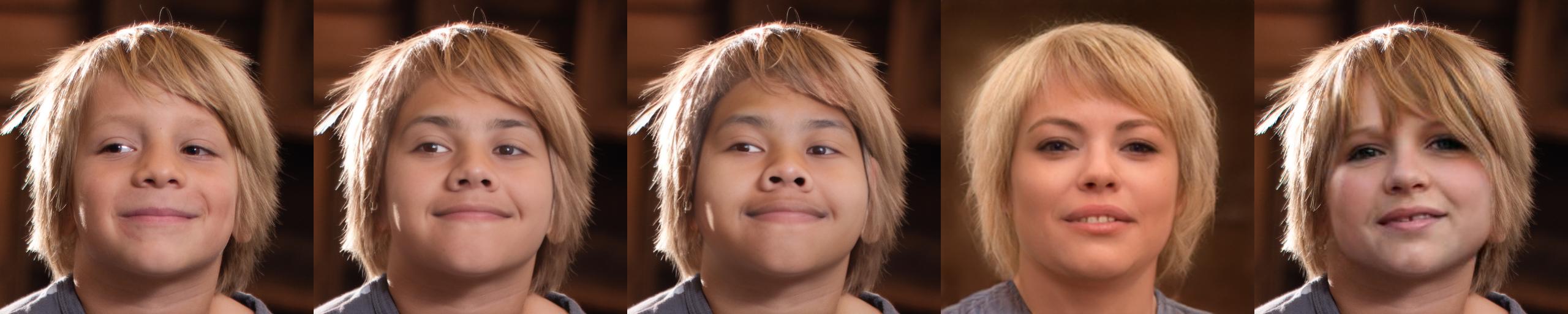}} \\
    Original & Ours \((d=1.2)\) & Ours \((d=1.4)\) & RiDDLE ~\cite{li2023riddle} & DP2~\cite{hukkelaas2023deepprivacy2} \\
  \end{tabular}
  \caption{Qualitative results for the face anonymization task for the FFHQ~\cite{karras2019style} test set.}
  \label{fig:anon_qual_ffhq}
\end{figure}

\paragraph{Qualitative Comparison.}
\Cref{fig:anon_qual_cele,fig:anon_qual_ffhq} present qualitative comparison results for anonymization tasks on the CelebA-HQ~\cite{karras2017progressive} and FFHQ~\cite{karras2019style} test sets, respectively. For face swapping tasks, \cref{fig:swap_qual} showcases two representative examples. Additional results are available in the supplementary material.

From \cref{fig:anon_qual_cele,fig:anon_qual_ffhq}, we observe that DP2~\cite{hukkelaas2023deepprivacy2} sometimes produces artifacts where the anonymized face does not align correctly with the position or orientation of the original face in the image. This issue arises because DP2~\cite{hukkelaas2023deepprivacy2} approaches anonymization as an image inpainting task. It first detects and crops the face region from the input photo, then applies a predicted mask over the region to be anonymized. An inpainting generator is then used to fill in these masked area with an anonymized face. However, if the mask inaccurately removes parts of the image, it can disrupt the inpainting process, leading to misaligned or distorted results. Our method overcomes these limitations of inpainting-based approaches by generating the entire image from a noise map, avoiding dependency on masks.

We also note that FALCO~\cite{barattin2023attribute} does not preserve background details because its design does not include background elements in its loss functions. Although FALCO~\cite{barattin2023attribute} incorporates facial attribute preservation loss, it struggles with maintaining certain facial features, such as eye direction, because it relies on finding similarity within the FaRL~\cite{zheng2022general} feature space, which does not encode eye gaze information. RiDDLE~\cite{li2023riddle} attempts to preserve image quality and similarity at the perceptual feature level by using a perceptual loss~\cite{zhang2018unreasonable}, but it still fails to accurately replicate specific details like eye direction, clothing, and background elements from the original image. In contrast, our method effectively modifies identity-related facial features while preserving non-identity-related details, thanks to its face-swapping approach and the advantages of its model architecture.

\begin{figure}
  \scriptsize
  \centering
  \begin{tabularx}{\columnwidth}{@{}X@{}X@{}X@{}X@{}X@{}X@{}}
    \multicolumn{6}{@{}c@{}}{\includegraphics[width=\linewidth]{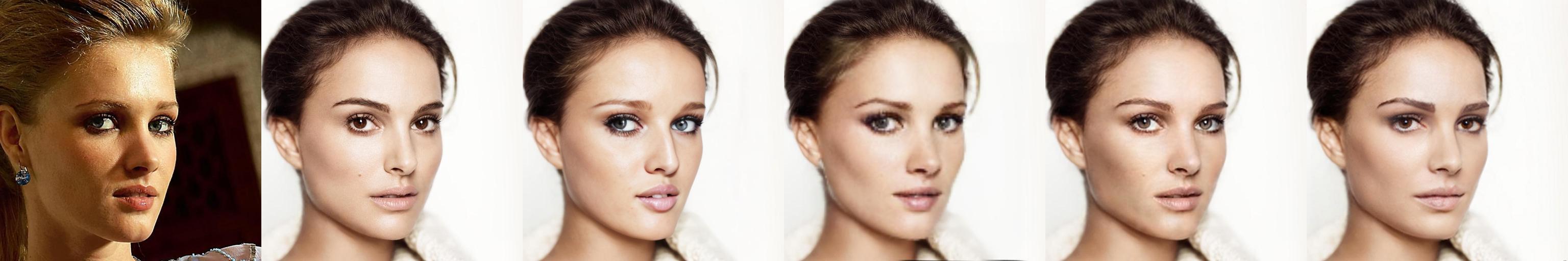}} \\
    \multicolumn{6}{@{}c@{}}{\includegraphics[width=\linewidth]{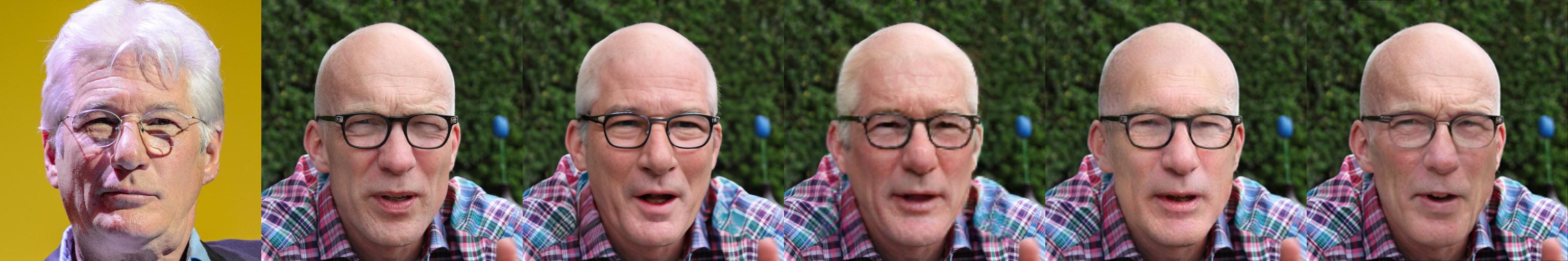}} \\
    \centering Source & \centering Driving & \centering Ours & \centering InSwapper ~\cite{Guo_InsightFace_Swapper} & \centering BlendFace ~\cite{shiohara2023blendface} & \centering DiffSwap ~\cite{zhao2023diffswap} \\
  \end{tabularx}
  \caption{Qualitative results for the face swapping task for the CelebA-HQ~\cite{karras2017progressive} test set in the upper row and the FFHQ~\cite{karras2019style} test set in the lower row.}
  \label{fig:swap_qual}
\end{figure}

\subsection{Ablation Study}

\begin{table}[ht]
  \footnotesize
  \begin{center}
    \begin{NiceTabular}{@{}c@{}}
      \begin{tabular}{l|cc|cc}
        & \multicolumn{4}{c}{Identity Distance} \\
        \cline{2-5}
        & \multicolumn{2}{c|}{Re-ID $\downarrow$} & \multicolumn{2}{c}{Shape $\uparrow$} \\
        & CelebA-HQ & FFHQ & CelebA-HQ & FFHQ \\
        \hline
        \sout{embeds} \tabularnote{Ours without modifying intermediate image embeddings} & 0.378 & 0.309 & 15.756 & 18.881 \\
        \sout{states} \tabularnote{Ours without modifying ReferenceNet states} & 0.288 & 0.545 & 21.342 & 16.566 \\
        \sout{uncond states} \tabularnote{Ours without including unconditional ReferenceNet states} & 0.159 & 0.243 & 17.559 & 18.867 \\
        Ours & \textbf{0.008} & \textbf{0.039} & \textbf{53.244} & \textbf{41.695} \\
      \end{tabular}
    \end{NiceTabular}
  \end{center}
  \caption{Ablation analysis of identity anonymization performance on the CelebA-HQ~\cite{karras2017progressive} and FFHQ~\cite{karras2019style} test sets, with the best results highlighted in bold.}
  \label{tab:ablation}
\end{table}

We conduct an ablation study on our anonymization approach, focusing on three key design elements related to \cref{eq:embedding,eq:state}: (1) unmodified intermediate image embeddings from the image encoder, (2) unmodified states of the source ReferenceNet model, and (3) modification limited to intermediate image embeddings from the image encoder and conditional states of the source ReferenceNet model, excluding its unconditional states.

\Cref{tab:ablation} presents the re-identification performance and face shape distance for our full method and each individual design choice. Our analysis reveals that: (1) modifying only the intermediate image embeddings or only the ReferenceNet states is not enough to improve re-identification performance or increase face shape distinctiveness. (2) Changing both the intermediate image embeddings and the conditional states of the source ReferenceNet model, without including its unconditional states, also fails to achieve significant improvements. (3) The key to substantially enhancing re-identification performance and creating less similar face shapes lies in a combined approach---modifying both the intermediate image embeddings and the conditional states of the source ReferenceNet model, while also incorporating its unconditional states.

The last row of \cref{tab:ablation}, representing our full method, demonstrates the effectiveness of this comprehensive approach.

\section{Conclusion}

We have introduced our approach leveraging diffusion models for face anonymization. Our framework eliminates the need for facial keypoints and masks and relies solely on a reconstruction loss, while still generating images with detailed fine-grained features. Our results show that this method effectively anonymizes faces, preserves attributes, and produces high-quality images. Additionally, our model can use an extra facial image input to perform face swapping tasks, demonstrating its versatility and potential for various facial image processing applications.

{\small
\bibliographystyle{ieee_fullname}
\bibliography{references}
}

\clearpage
\twocolumn[{
  \renewcommand\twocolumn[1][]{#1}
  \centering
  \Large
  \textbf{Face Anonymization Made Simple}\\
  \vspace{1.5em}Supplementary Material \\
  \vspace{1.0em}
}]

\section*{Anonymization Degree vs. Identity Distance}

\begin{figure}[b]
  \centering
  \includegraphics[width=0.8\linewidth]{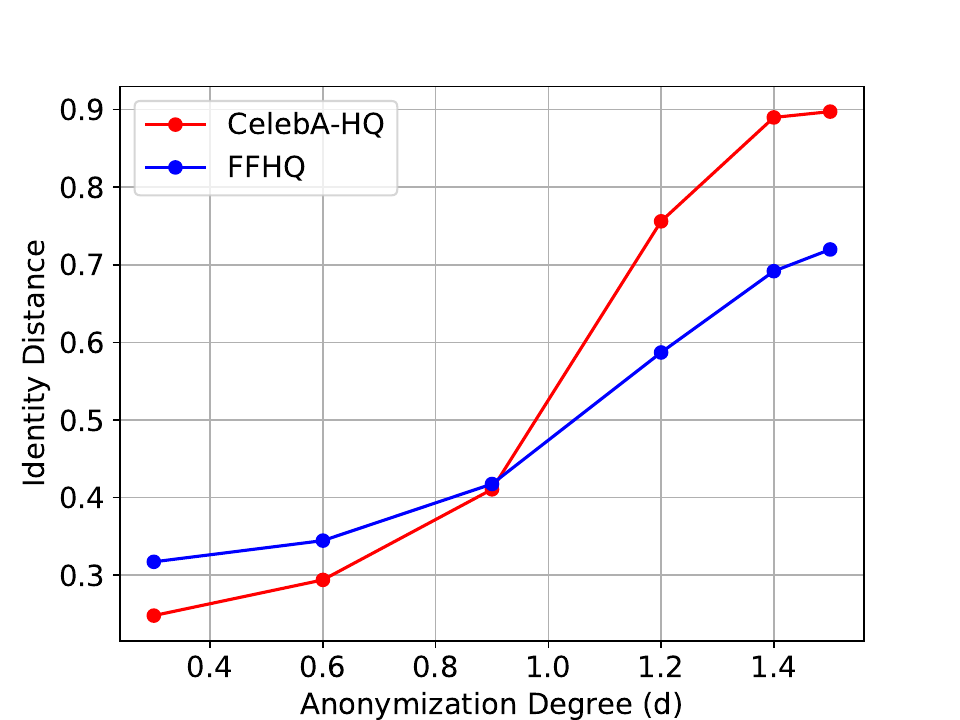}
  \caption{Relationship between degree of anonymization and identity distance.}
  \label{fig:anon_degree_vs_id_dist}
\end{figure}

Our anonymization approach features a single adjustable parameter, $d$, which controls the degree of anonymization. \Cref{fig:anon_degree_vs_id_dist} illustrates how the identity cosine distance, measured using the FaceNet~\cite{schroff2015facenet} recognition model, changes with varying degrees of anonymization. This analysis was conducted using dozens of identities and seeds from our CelebA-HQ~\cite{karras2017progressive} and FFHQ~\cite{karras2019style} test sets. We selected 50 identities from each dataset. For each identity, we applied six different degrees of anonymization ($d$ values of 0.3, 0.6, 0.9, 1.2, 1.4, 1.5). For each anonymization degree, we created 10 variations using 10 different seed values. The plot displays the average cosine distance of these variations for each anonymization degree. It reveals a clear trend: as the degree of anonymization increases, the identity distance between the anonymized and original images grows wider.

\section*{Optimizing Face Anonymization}

\begin{figure}[b]
  \centering
  \includegraphics[width=0.8\linewidth]{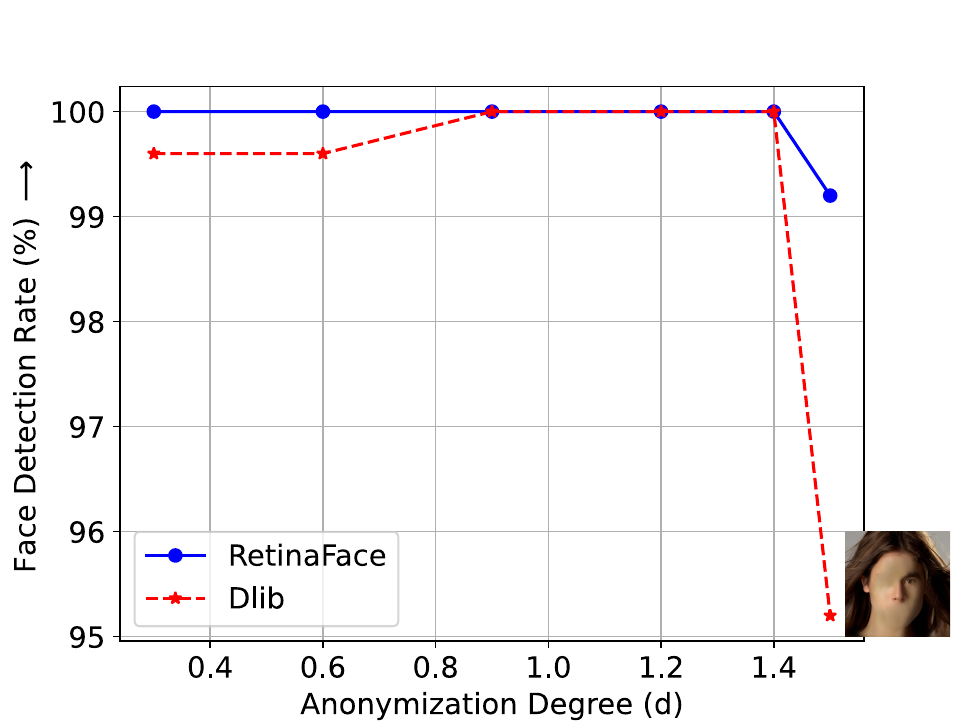}
  \caption{Face detection rates at various anonymization degrees.}
  \label{fig:detection_rates}
\end{figure}

Increasing the anonymization degree parameter, $d$, results in a greater divergence from the original identity, but beyond a certain range, it can hinder the model's ability to generate realistic faces. We followed prior research methodologies~\cite{barattin2023attribute,helou2023vera,hukkelaas2019deepprivacy} and used face detection to assess the validity of synthesized faces. We applied six different anonymization levels ($d$ values of 0.3, 0.6, 0.9, 1.2, 1.4, and 1.5) to 250 facial images from our CelebA-HQ~\cite{karras2017progressive} test set and evaluated the detection rate with two face detectors, RetinaFace~\cite{deng2020retinaface} and Dlib~\cite{king2009dlib}. As shown in \cref{fig:detection_rates}, when $d$ reaches 1.5, the face detectors begin to flag invalid faces among the 250 generated, indicating that $d$ values above 1.5 are unsuitable for maintaining realistic outputs. \Cref{fig:detection_rates} also includes an example illustrating the unrealistic faces generated by our model at higher $d$ values.

\begin{figure}
  \centering
  \begin{subfigure}{0.48\linewidth}
    \centering
    \includegraphics[width=1.0\linewidth]{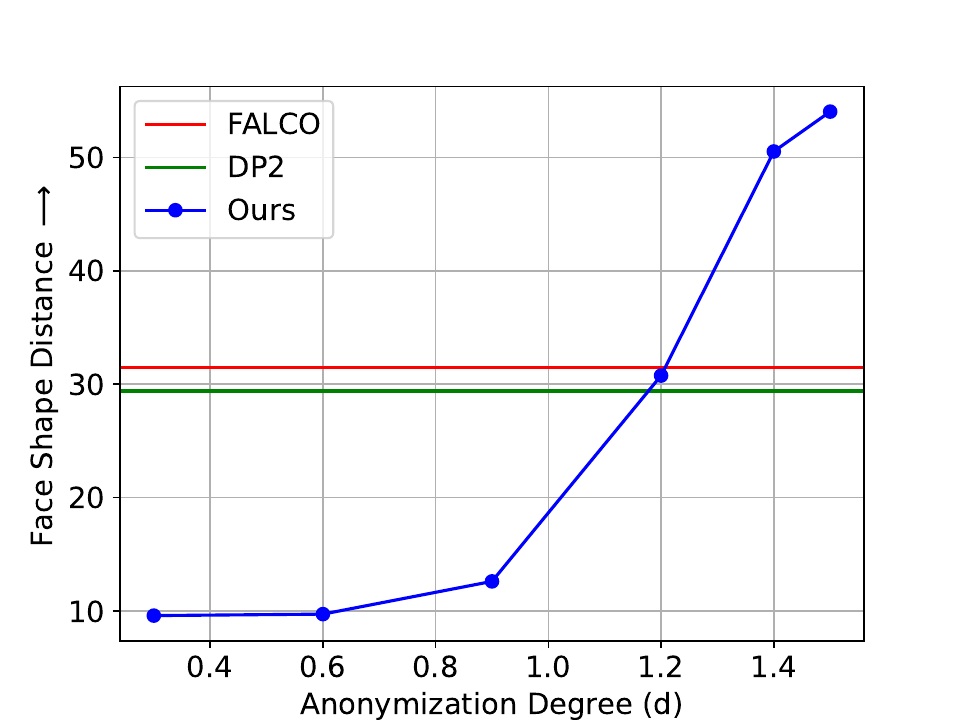}
    \caption{}
    \label{fig:features-a}
  \end{subfigure}
  \hfill
  \begin{subfigure}{0.48\linewidth}
    \centering
    \includegraphics[width=1.0\linewidth]{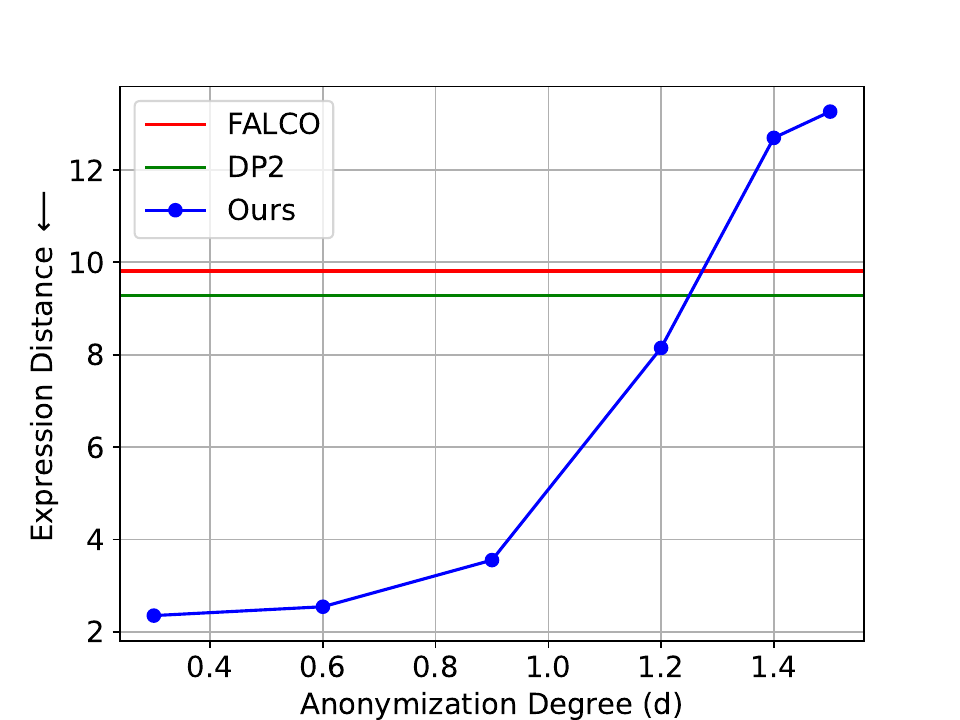}
    \caption{}
    \label{fig:features-b}
  \end{subfigure}
  \\
  \begin{subfigure}{0.48\linewidth}
    \centering
    \includegraphics[width=1.0\linewidth]{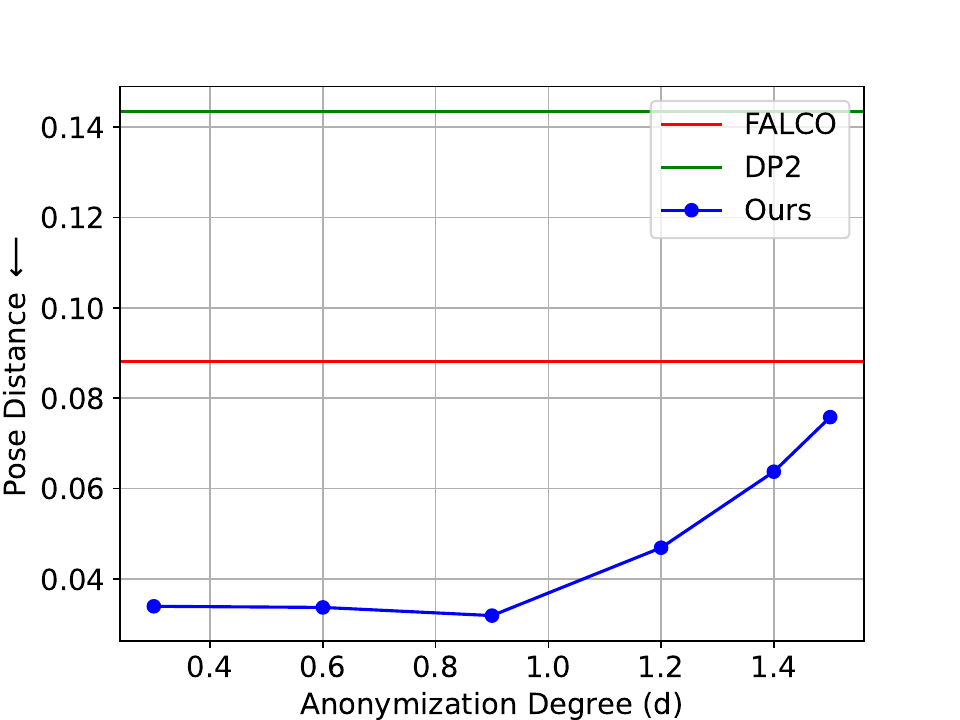}
    \caption{}
    \label{fig:features-c}
  \end{subfigure}
  \hfill
  \begin{subfigure}{0.48\linewidth}
    \centering
    \includegraphics[width=1.0\linewidth]{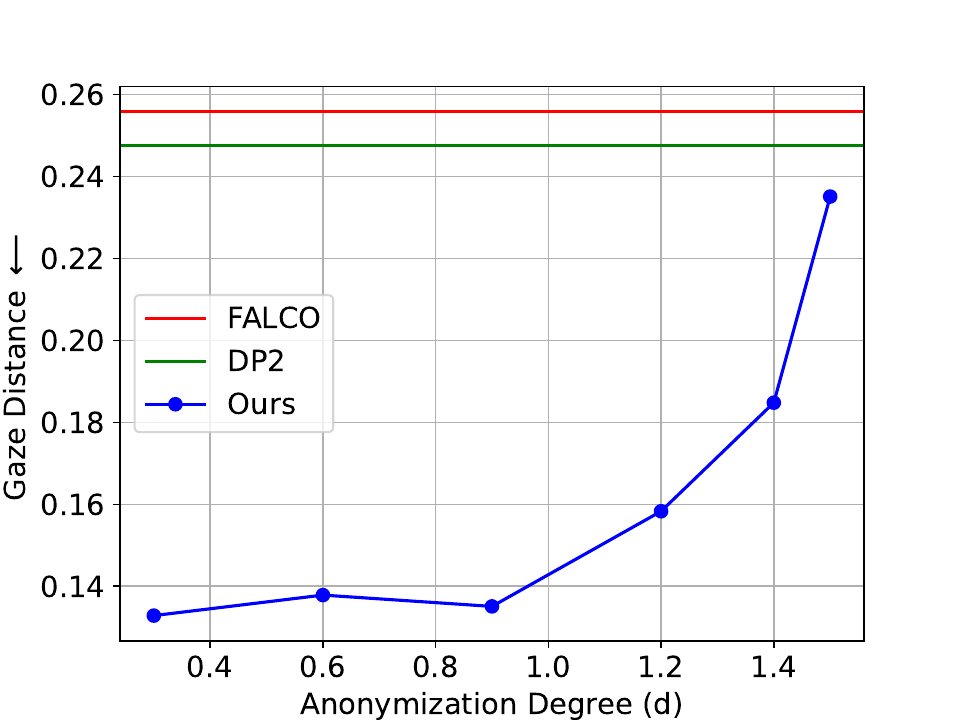}
    \caption{}
    \label{fig:features-d}
  \end{subfigure}
  \\
  \caption{Changes in facial features at different anonymization degrees.}
  \label{fig:facial_features}
\end{figure}

\begin{table*}[ht!]
  \resizebox{\textwidth}{!}{
    \begin{tabular}{l|cc|cc|cc|cc|cc|cc}
      & \multicolumn{4}{c|}{Identity Distance} & \multicolumn{6}{c|}{Attribute Distance} & \multicolumn{2}{c}{Image Quality} \\
      \cline{2-13}
      & \multicolumn{2}{c|}{Re-ID $\uparrow$} & \multicolumn{2}{c|}{Shape $\downarrow$} & \multicolumn{2}{c|}{Pose $\downarrow$} & \multicolumn{2}{c|}{Gaze $\downarrow$} & \multicolumn{2}{c|}{Expression $\downarrow$} & \multicolumn{2}{c}{Face IQA $\uparrow$} \\
      & CelebA-HQ & FFHQ & CelebA-HQ & FFHQ & CelebA-HQ & FFHQ & CelebA-HQ & FFHQ & CelebA-HQ & FFHQ & CelebA-HQ & FFHQ \\
      \hline
      DiffSwap~\cite{zhao2023diffswap} & 0.114 & 0.162 & 16.590 & 17.914 & \underline{0.034} & \underline{0.041} & 0.151 & 0.164 & \textbf{2.682} & \textbf{3.195} & \underline{0.549} & \underline{0.542} \\
      BlendFace~\cite{shiohara2023blendface} & \underline{0.693} & \underline{0.642} & \underline{13.234} & \underline{16.497} & \textbf{0.028} & \textbf{0.036} & \textbf{0.120} & \textbf{0.148} & \underline{3.170} & \underline{4.102} & 0.527 & 0.511 \\
      InSwapper~\cite{Guo_InsightFace_Swapper} & \textbf{0.871} & \textbf{0.830} & \textbf{11.558} & \textbf{14.300} & 0.035 & 0.042 & 0.158 & 0.177 & 4.197 & 4.872 & 0.364 & 0.371 \\
      Ours & 0.566 & 0.310 & 17.211 & 22.312 & 0.036 & 0.043 & \underline{0.139} & \underline{0.149} & 4.067 & 4.745 & \textbf{0.728} & \textbf{0.720} \\
    \end{tabular}
  }
  \caption{Quantitative results on the task of face swapping for CelebA-HQ~\cite{karras2017progressive} and FFHQ~\cite{karras2019style} test sets, with the best results highlighted in bold and the second-best results underlined.}
  \label{tab:swap_quan}
\end{table*}

While higher $d$ values create more distinct face shapes for anonymization, they compromise the preservation of non-identity related facial attributes. We again applied the same six levels of anonymization to 250 facial images from our CelebA-HQ~\cite{karras2017progressive} test set and measured the attribute distances for face shape, pose, gaze, and expression. \Cref{fig:features-a} reveals that higher $d$ values produce more distinctive face shapes, but this comes at the cost of preserving non-identity-related attributes. This trend is further illustrated in \cref{fig:features-b,fig:features-c,fig:features-d}, aligning with our expectations.

We also evaluated the attribute distance performance of two state-of-the-art anonymization methods, FALCO~\cite{barattin2023attribute} and DP2~\cite{hukkelaas2023deepprivacy2}, on the same 250 test images, presenting their results in ~\cref{fig:facial_features}. For the FALCO~\cite{barattin2023attribute} method, we set its identity loss margin value to 0. This configuration maximizes the identity difference between the anonymized result and the original image, but it compromises the preservation of non-identity facial attributes. \Cref{fig:features-a} reveals that when the $d$ value exceeds 1.2, our method begins to outperform these methods in producing more distinctive face shapes. Additionally, our method may continue to outperform them in preserving facial expressions until $d$ reaches approximately 1.3, as illustrated in \cref{fig:features-b}. Based on our empirical results, we recommend an optimal $d$ range of 1.2 to 1.3 to achieve the ideal balance between identity obfuscation and attribute preservation. Within this range, our method also demonstrates superior performance compared to current state-of-the-art anonymization techniques.

\section*{Dataset Preparation for Model Training}

For training our model, we used three datasets: CelebRef-HQ~\cite{li2022learning}, CelebA-HQ~\cite{karras2017progressive}, and FFHQ~\cite{karras2019style}. We used all 10,555 images from CelebRef-HQ~\cite{li2022learning}, which contains 1,005 identities with multiple images per identity showing varied expressions and angles. For CelebA-HQ~\cite{karras2017progressive} and FFHQ~\cite{karras2019style}, we employed face recognition to identify same-person images, selecting 6,203 images (2,506 identities) from CelebA-HQ~\cite{karras2017progressive} and 7,816 images (2,887 identities) from FFHQ~\cite{karras2019style}. For each identity, we randomly chose two images: one as the source and one as ground truth. We then created a synthesized driving image by using a state-of-the-art face swapping model~\cite{Guo_InsightFace_Swapper} to replace the face in the ground truth image with another person's face. This process resulted in 153,414 source-driving image pairs for training: 49,518 from CelebA-HQ~\cite{karras2017progressive}, 42,188 from CelebRef-HQ~\cite{li2022learning}, and 61,708 from FFHQ~\cite{karras2019style}.

\section*{Qualitative Results of Anonymization}

\begin{figure*}
  \footnotesize
  \centering
  \begin{tabular}{*{5}{>{\centering\arraybackslash}m{\dimexpr.18\linewidth-2\tabcolsep}}}
    \multicolumn{5}{@{}c@{}}{\includegraphics[width=.9\linewidth]{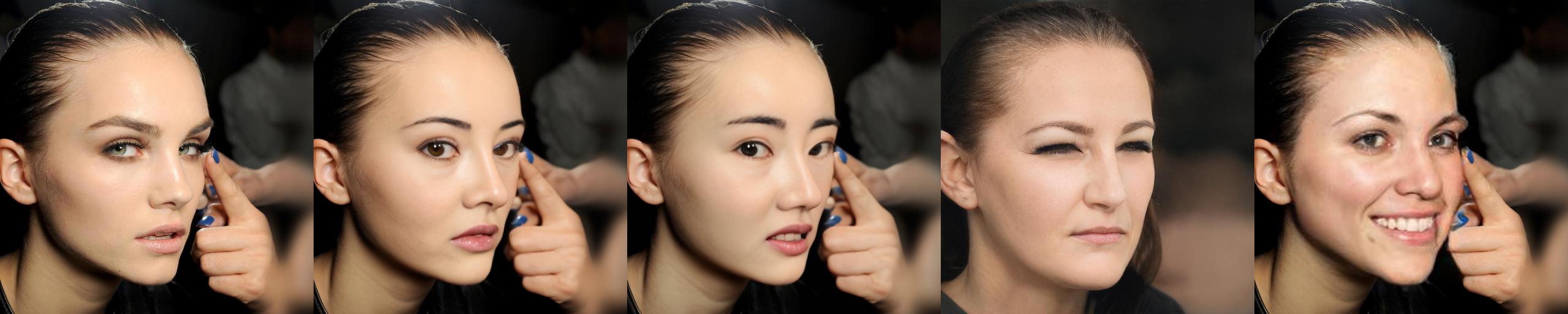}} \\
    \multicolumn{5}{@{}c@{}}{\includegraphics[width=.9\linewidth]{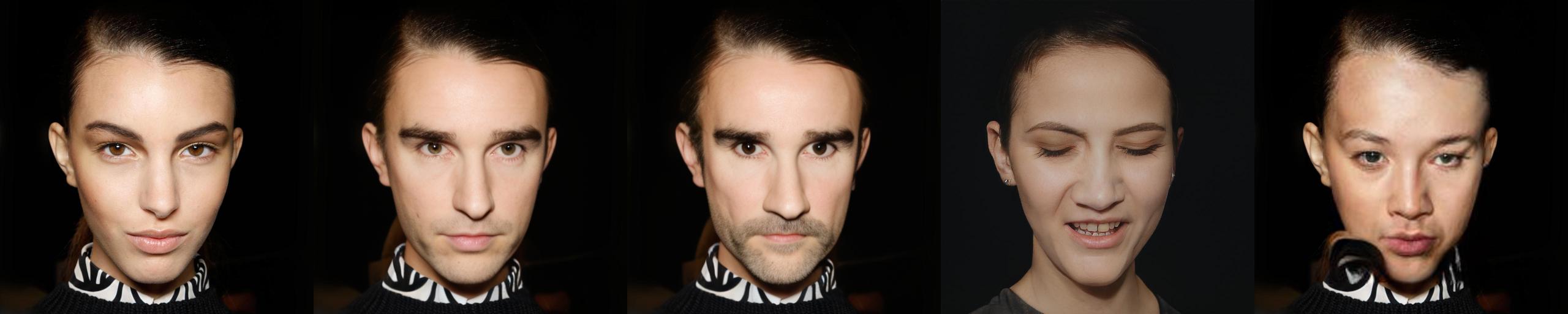}} \\
    \multicolumn{5}{@{}c@{}}{\includegraphics[width=.9\linewidth]{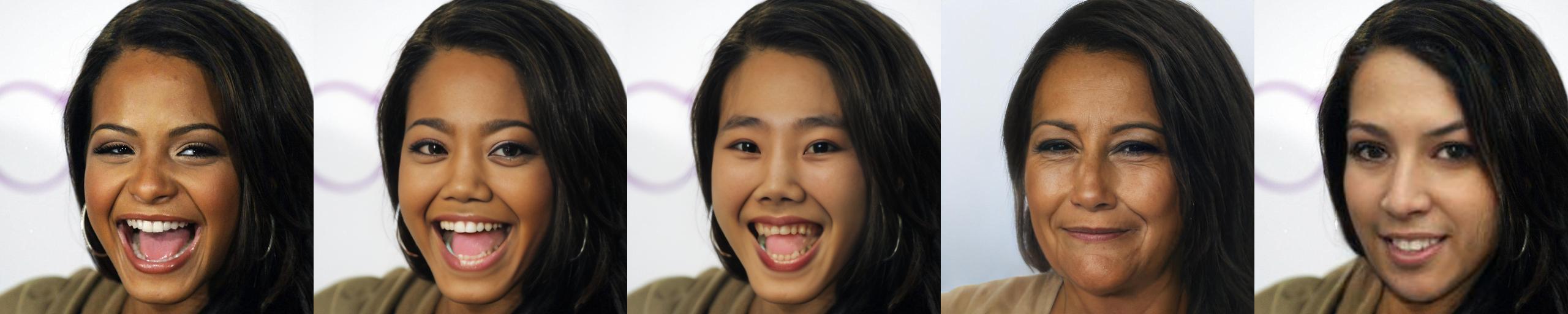}} \\
    \multicolumn{5}{@{}c@{}}{\includegraphics[width=.9\linewidth]{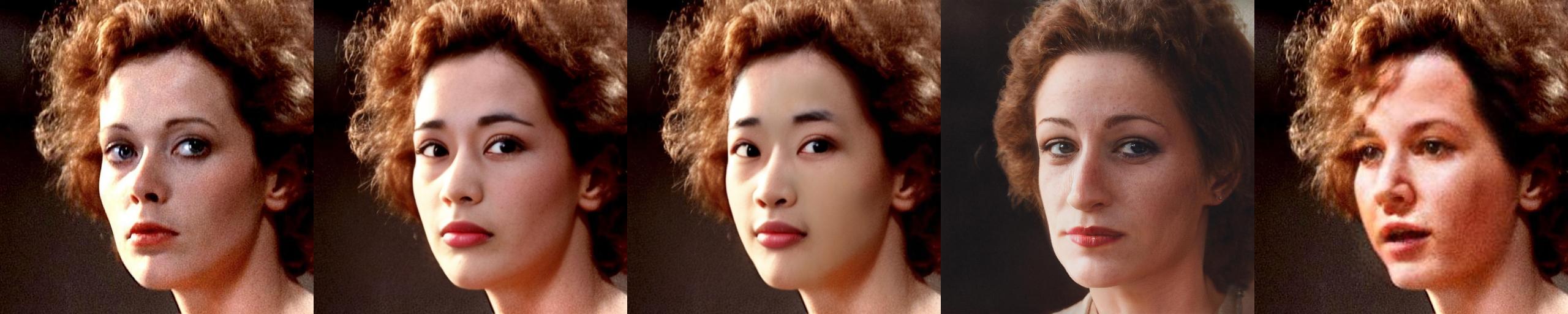}} \\
    \multicolumn{5}{@{}c@{}}{\includegraphics[width=.9\linewidth]{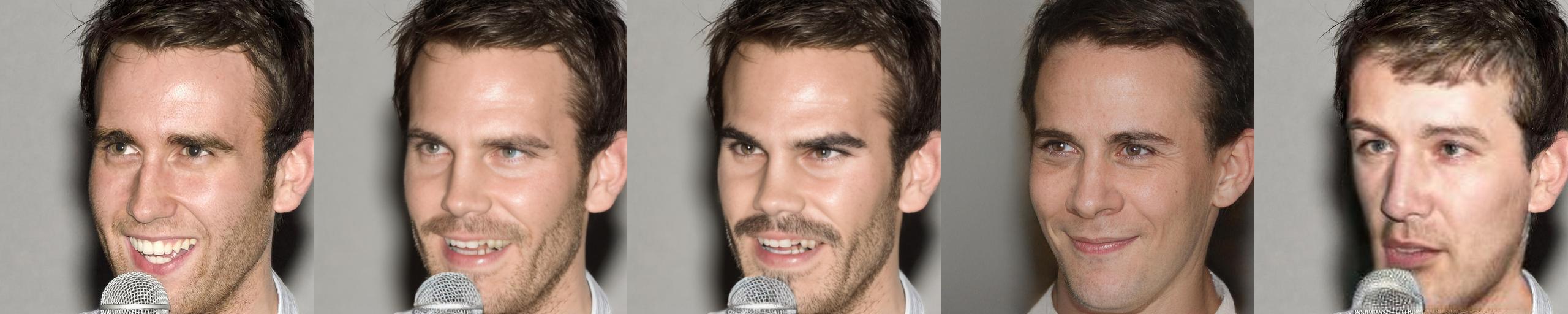}} \\
    \multicolumn{5}{@{}c@{}}{\includegraphics[width=.9\linewidth]{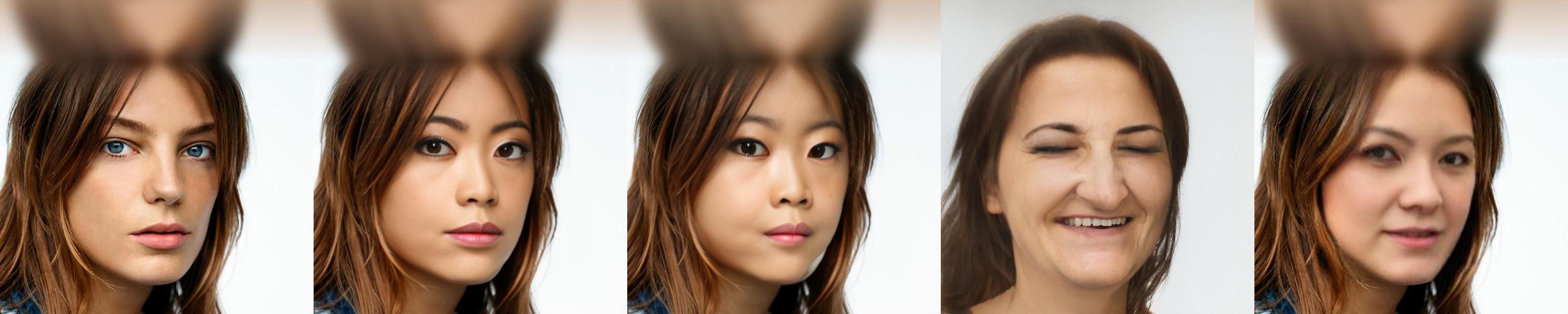}} \\
    Original & Ours \((d=1.2)\) & Ours \((d=1.4)\) & FALCO~\cite{barattin2023attribute} & DP2~\cite{hukkelaas2023deepprivacy2} \\
  \end{tabular}
  \caption{Qualitative results on the task of face anonymization for CelebA-HQ~\cite{karras2017progressive} test set.}
  \label{fig:anon_qual_cele1}
\end{figure*}

\begin{figure*}
  \footnotesize
  \centering
  \begin{tabular}{*{5}{>{\centering\arraybackslash}m{\dimexpr.18\linewidth-2\tabcolsep}}}
    \multicolumn{5}{@{}c@{}}{\includegraphics[width=.9\linewidth]{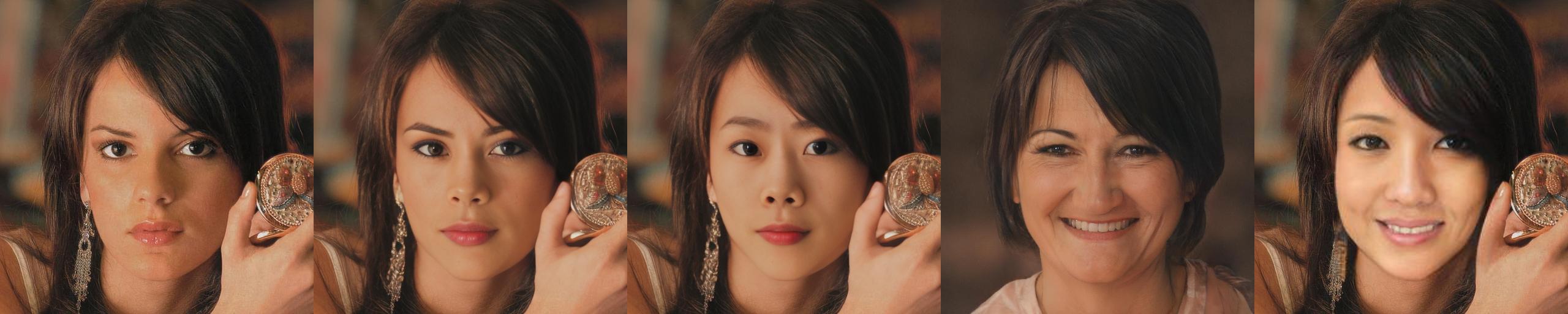}} \\
    \multicolumn{5}{@{}c@{}}{\includegraphics[width=.9\linewidth]{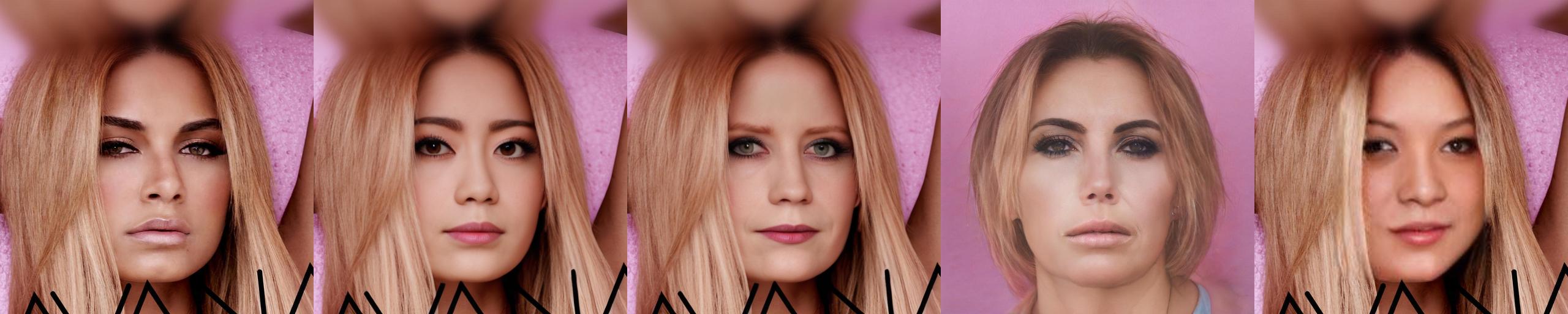}} \\
    \multicolumn{5}{@{}c@{}}{\includegraphics[width=.9\linewidth]{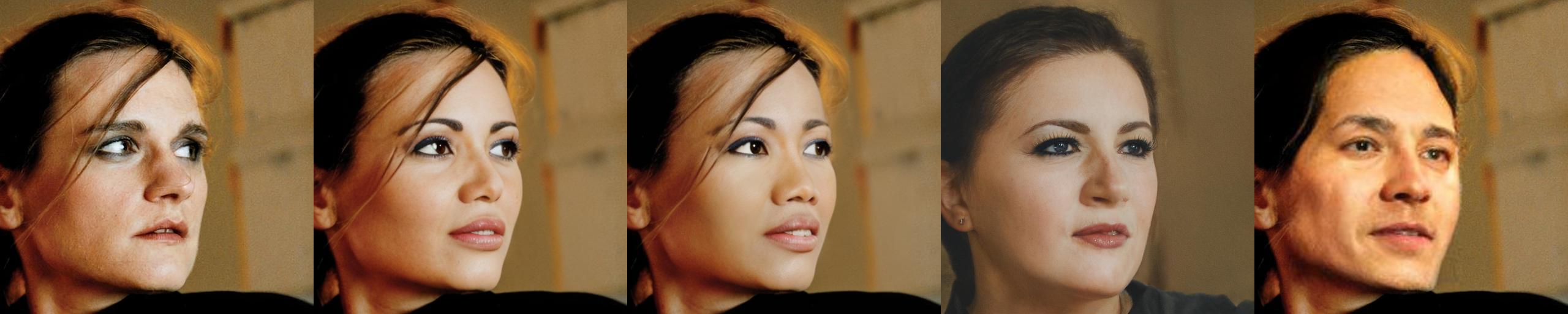}} \\
    \multicolumn{5}{@{}c@{}}{\includegraphics[width=.9\linewidth]{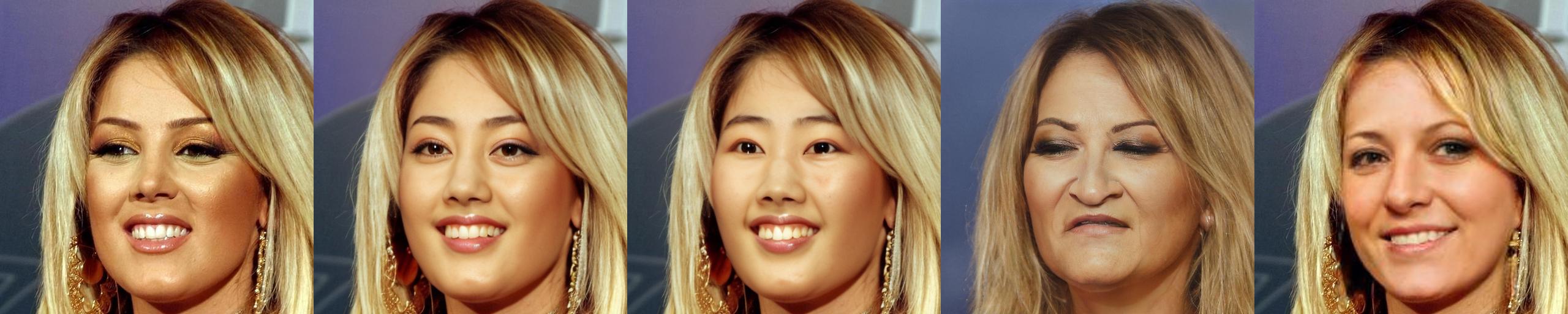}} \\
    \multicolumn{5}{@{}c@{}}{\includegraphics[width=.9\linewidth]{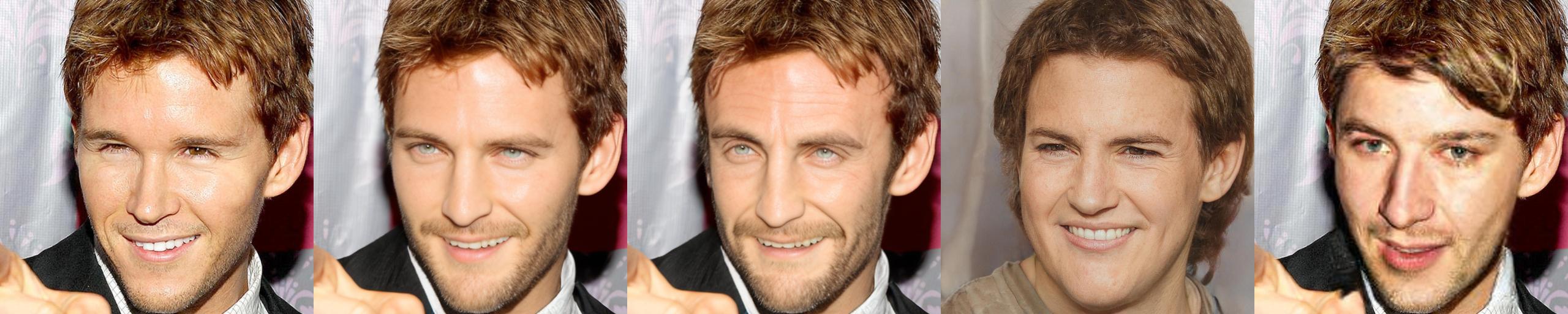}} \\
    \multicolumn{5}{@{}c@{}}{\includegraphics[width=.9\linewidth]{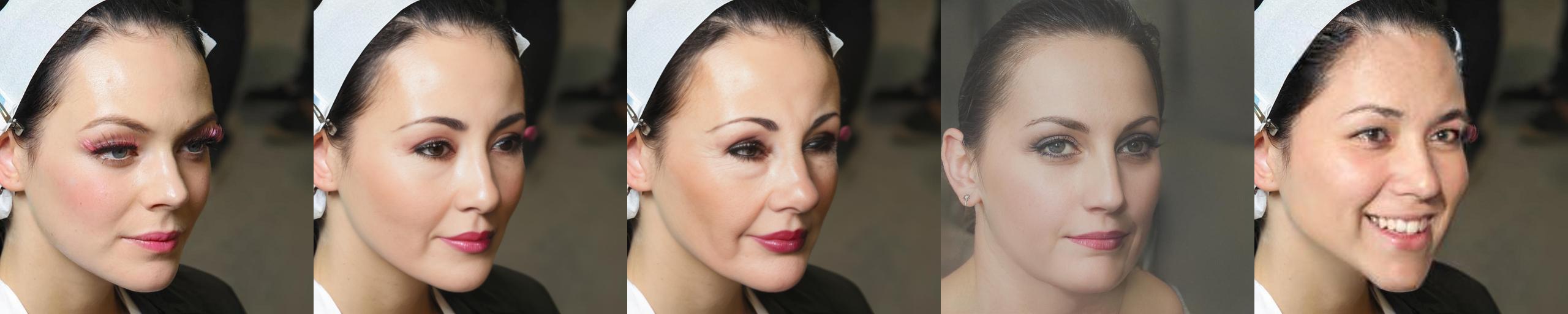}} \\
    Original & Ours \((d=1.2)\) & Ours \((d=1.4)\) & FALCO~\cite{barattin2023attribute} & DP2~\cite{hukkelaas2023deepprivacy2} \\
  \end{tabular}
  \caption{Qualitative results on the task of face anonymization for CelebA-HQ~\cite{karras2017progressive} test set.}
  \label{fig:anon_qual_cele2}
\end{figure*}

\begin{figure*}
  \footnotesize
  \centering
  \begin{tabular}{*{5}{>{\centering\arraybackslash}m{\dimexpr.18\linewidth-2\tabcolsep}}}
    \multicolumn{5}{@{}c@{}}{\includegraphics[width=.9\linewidth]{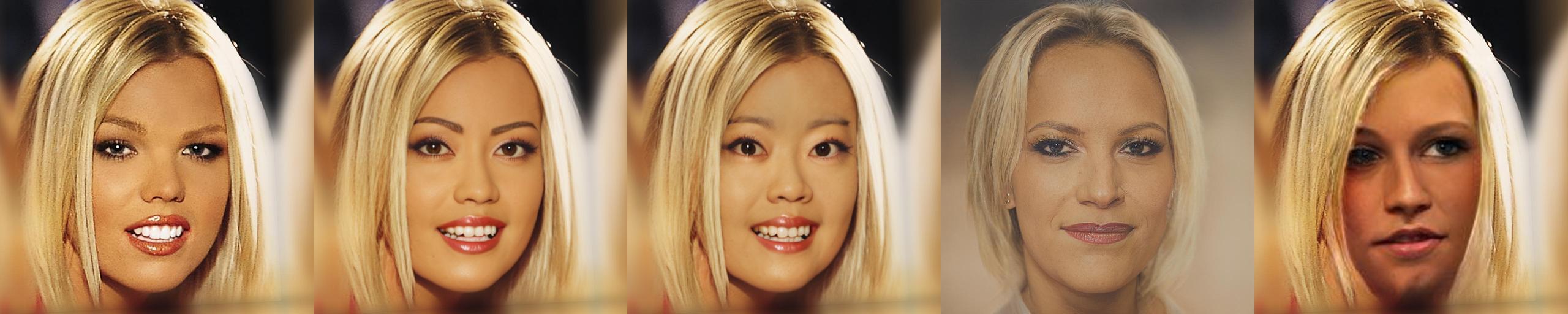}} \\
    \multicolumn{5}{@{}c@{}}{\includegraphics[width=.9\linewidth]{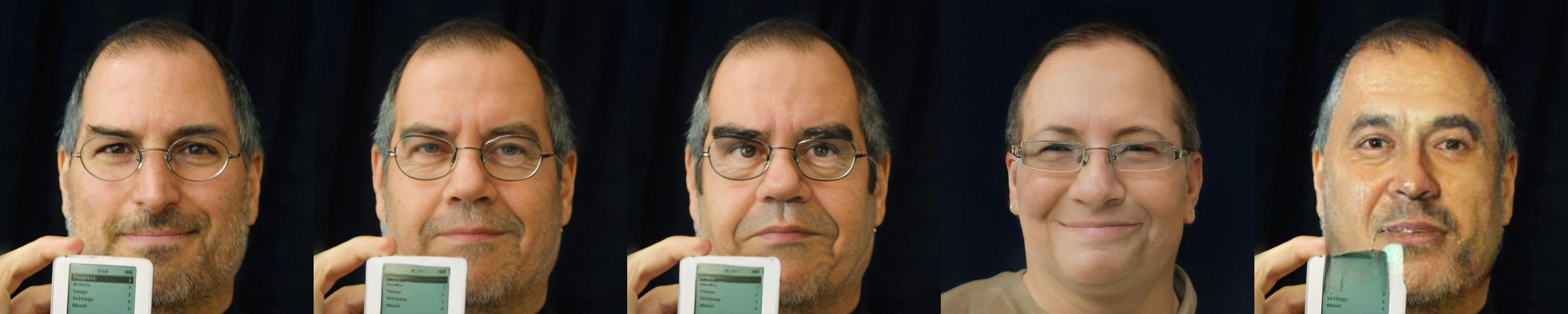}} \\
    \multicolumn{5}{@{}c@{}}{\includegraphics[width=.9\linewidth]{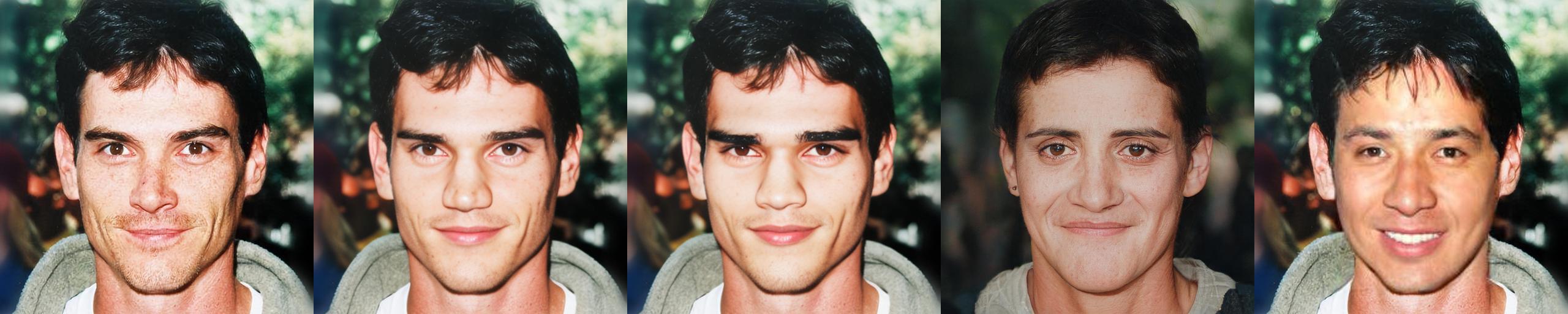}} \\
    \multicolumn{5}{@{}c@{}}{\includegraphics[width=.9\linewidth]{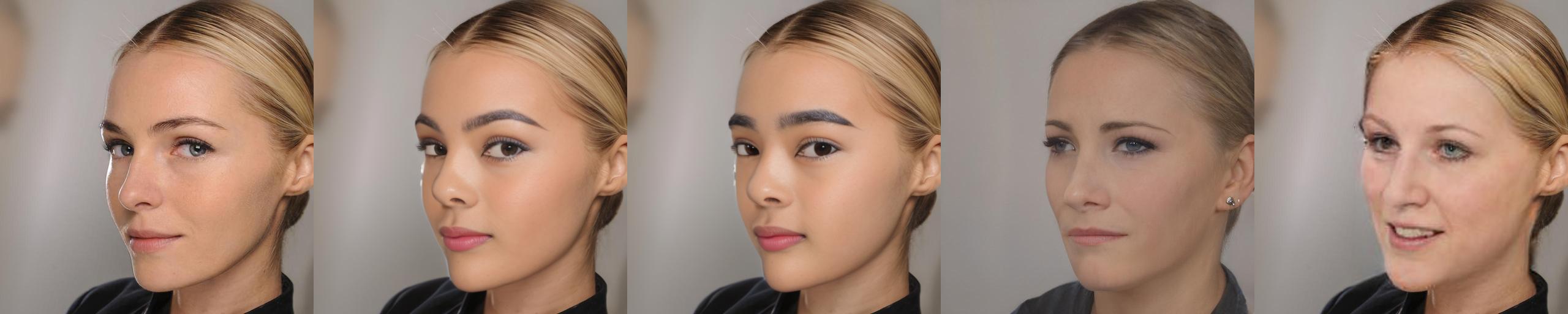}} \\
    \multicolumn{5}{@{}c@{}}{\includegraphics[width=.9\linewidth]{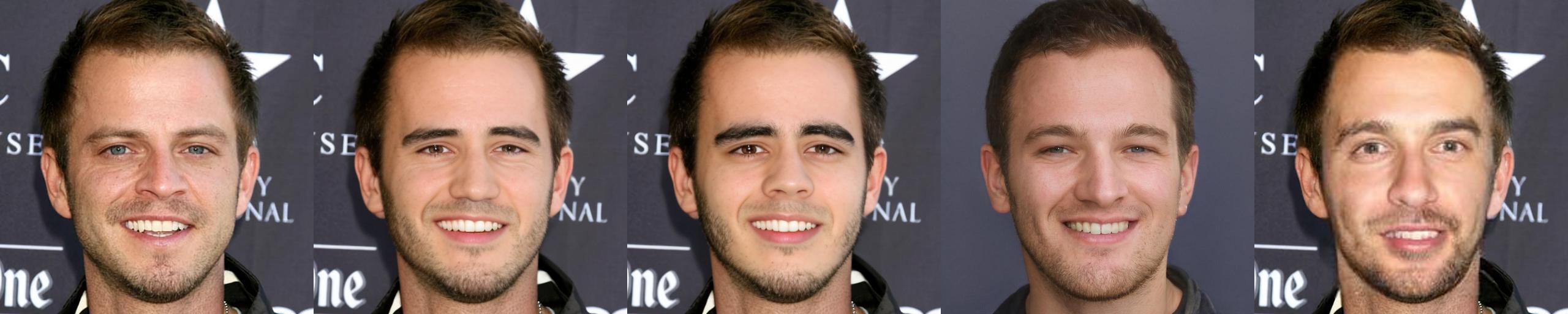}} \\
    \multicolumn{5}{@{}c@{}}{\includegraphics[width=.9\linewidth]{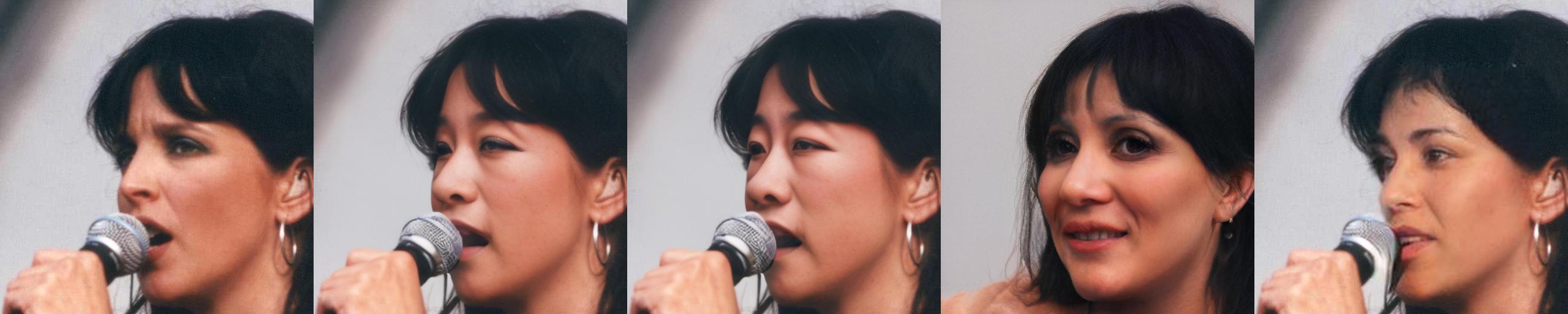}} \\
    Original & Ours \((d=1.2)\) & Ours \((d=1.4)\) & FALCO~\cite{barattin2023attribute} & DP2~\cite{hukkelaas2023deepprivacy2} \\
  \end{tabular}
  \caption{Qualitative results on the task of face anonymization for CelebA-HQ~\cite{karras2017progressive} test set.}
  \label{fig:anon_qual_cele3}
\end{figure*}

\begin{figure*}
  \footnotesize
  \centering
  \begin{tabular}{*{5}{>{\centering\arraybackslash}m{\dimexpr.18\linewidth-2\tabcolsep}}}
    \multicolumn{5}{@{}c@{}}{\includegraphics[width=.9\linewidth]{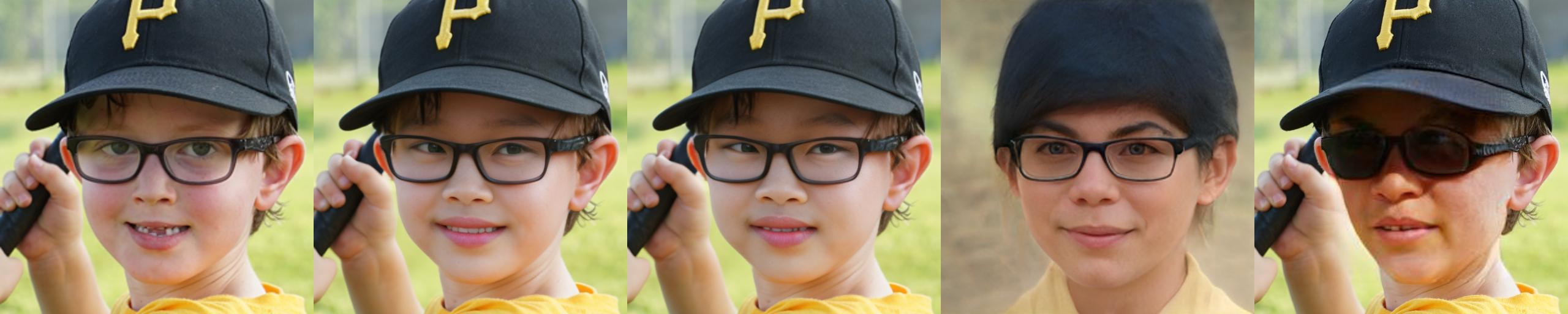}} \\
    \multicolumn{5}{@{}c@{}}{\includegraphics[width=.9\linewidth]{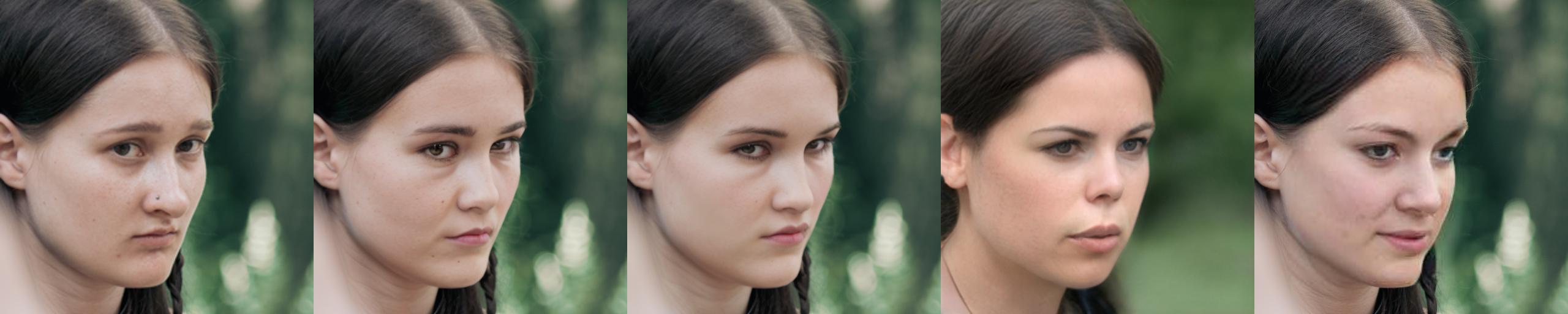}} \\
    \multicolumn{5}{@{}c@{}}{\includegraphics[width=.9\linewidth]{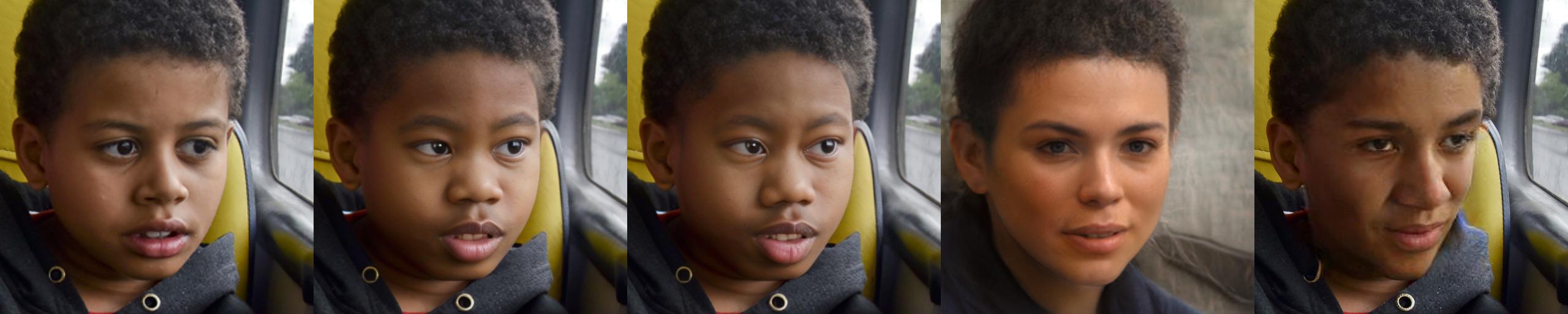}} \\
    \multicolumn{5}{@{}c@{}}{\includegraphics[width=.9\linewidth]{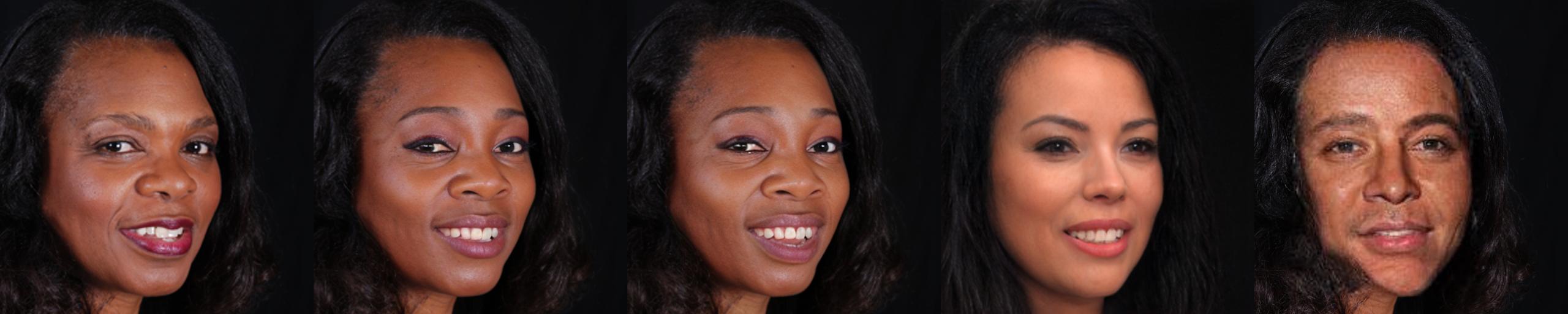}} \\
    \multicolumn{5}{@{}c@{}}{\includegraphics[width=.9\linewidth]{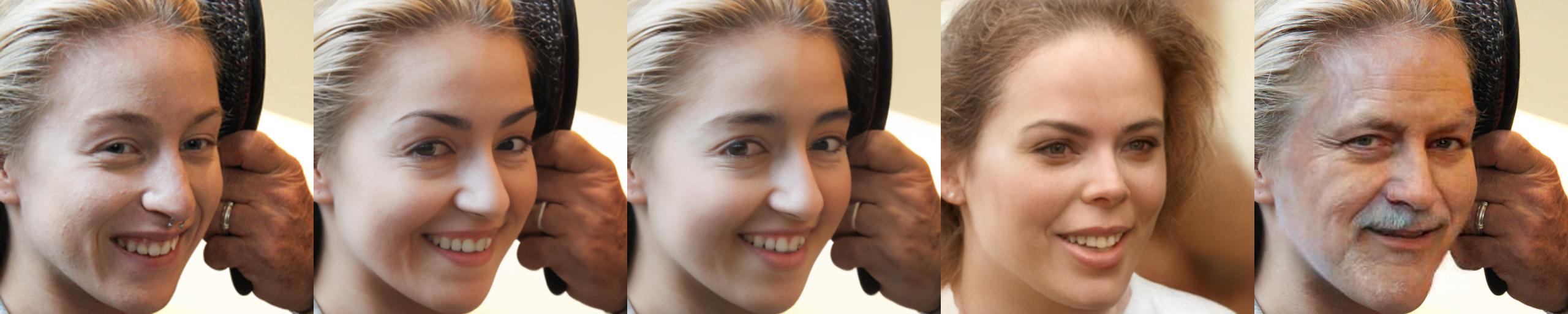}} \\
    \multicolumn{5}{@{}c@{}}{\includegraphics[width=.9\linewidth]{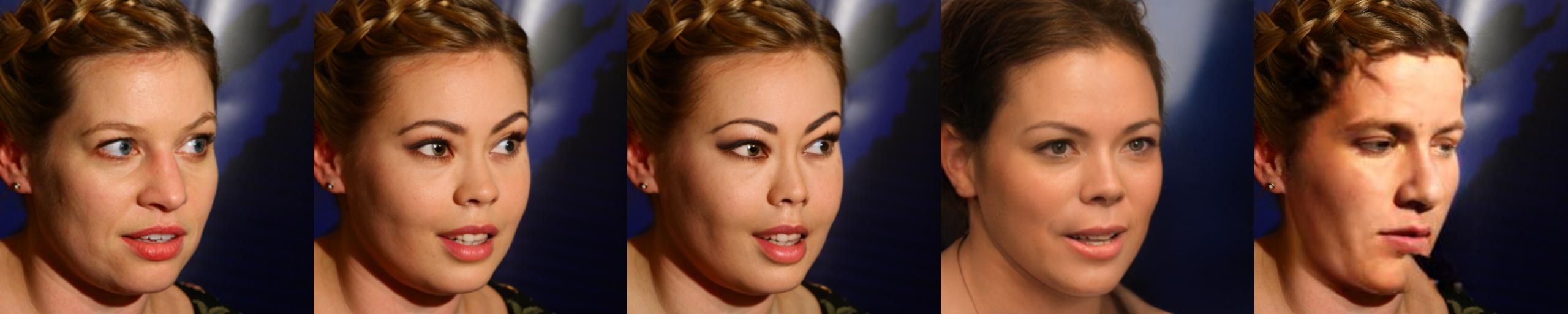}} \\
    Original & Ours \((d=1.2)\) & Ours \((d=1.4)\) & RiDDLE~\cite{li2023riddle} & DP2~\cite{hukkelaas2023deepprivacy2} \\
  \end{tabular}
  \caption{Qualitative results on the task of face anonymization for FFHQ~\cite{karras2019style} test set.}
  \label{fig:anon_qual_ffhq1}
\end{figure*}

\begin{figure*}
  \footnotesize
  \centering
  \begin{tabular}{*{5}{>{\centering\arraybackslash}m{\dimexpr.18\linewidth-2\tabcolsep}}}
    \multicolumn{5}{@{}c@{}}{\includegraphics[width=.9\linewidth]{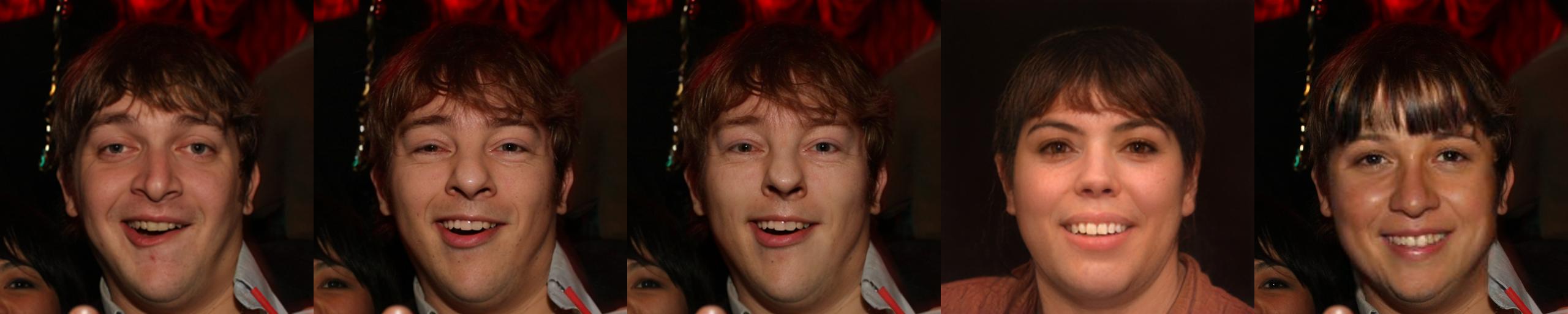}} \\
    \multicolumn{5}{@{}c@{}}{\includegraphics[width=.9\linewidth]{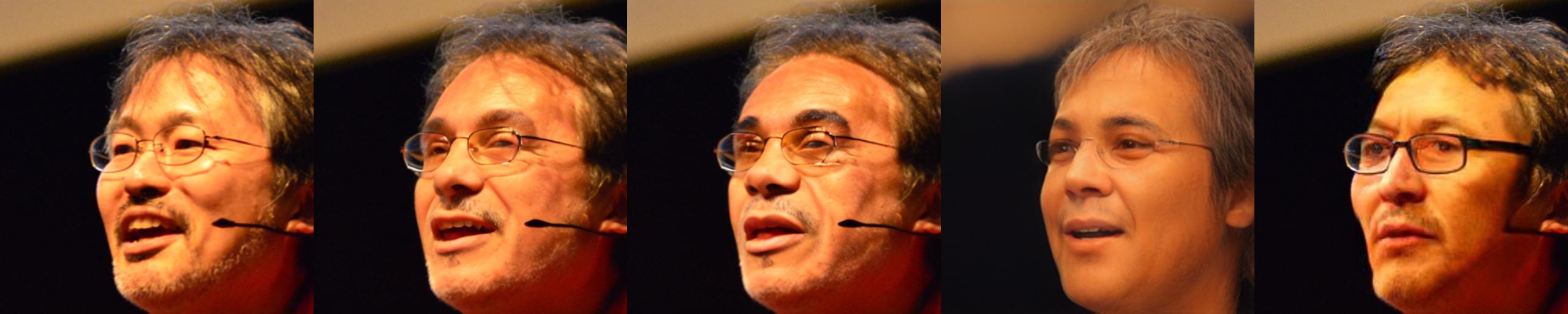}} \\
    \multicolumn{5}{@{}c@{}}{\includegraphics[width=.9\linewidth]{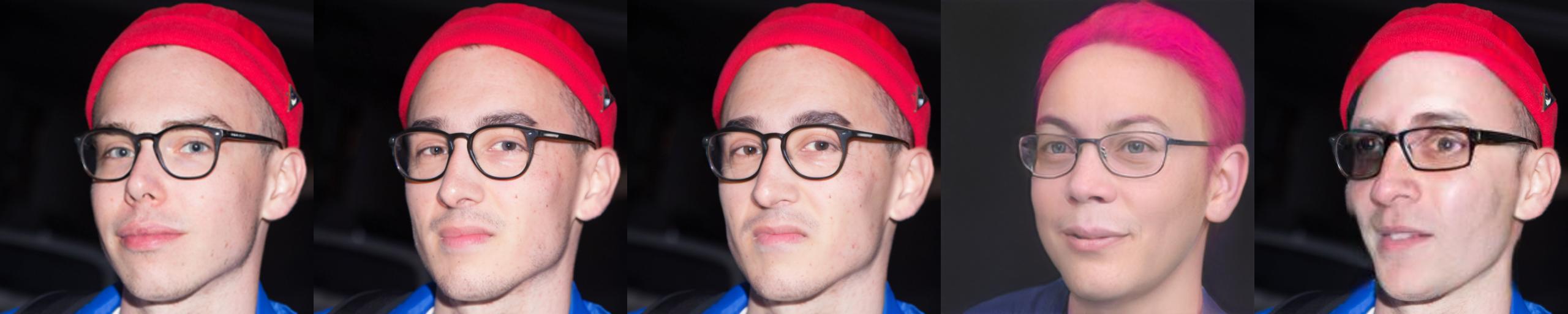}} \\
    \multicolumn{5}{@{}c@{}}{\includegraphics[width=.9\linewidth]{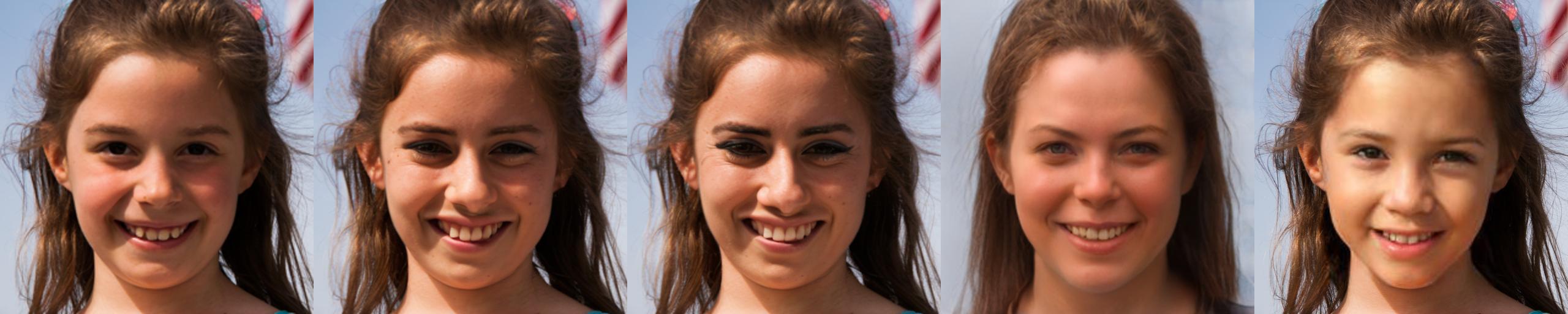}} \\
    \multicolumn{5}{@{}c@{}}{\includegraphics[width=.9\linewidth]{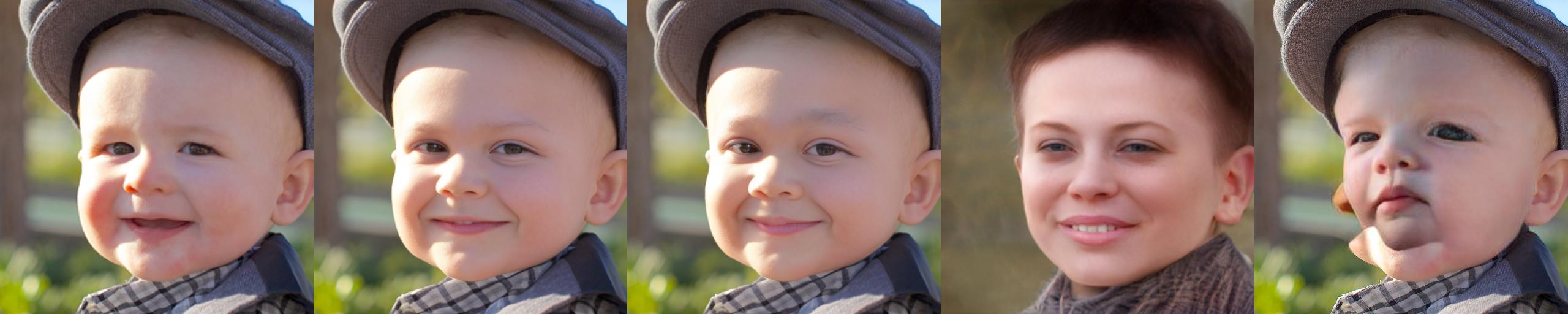}} \\
    \multicolumn{5}{@{}c@{}}{\includegraphics[width=.9\linewidth]{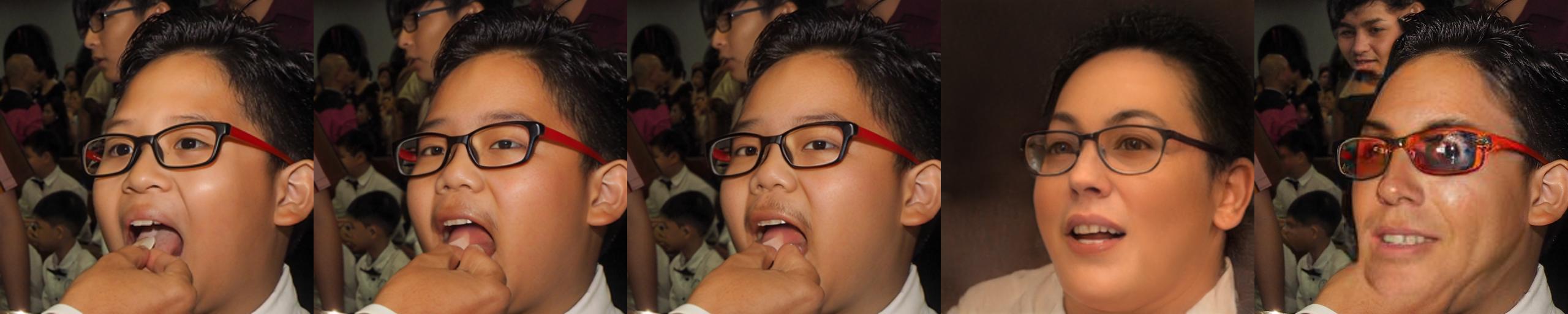}} \\
    Original & Ours \((d=1.2)\) & Ours \((d=1.4)\) & RiDDLE~\cite{li2023riddle} & DP2~\cite{hukkelaas2023deepprivacy2} \\
  \end{tabular}
  \caption{Qualitative results on the task of face anonymization for FFHQ~\cite{karras2019style} test set.}
  \label{fig:anon_qual_ffhq2}
\end{figure*}

\begin{figure*}
  \footnotesize
  \centering
  \begin{tabular}{*{5}{>{\centering\arraybackslash}m{\dimexpr.18\linewidth-2\tabcolsep}}}
    \multicolumn{5}{@{}c@{}}{\includegraphics[width=.9\linewidth]{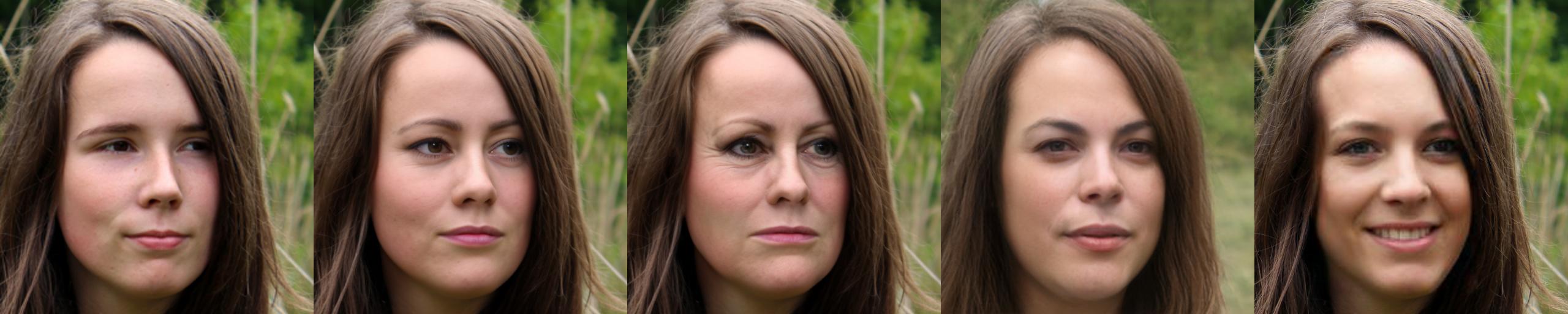}} \\
    \multicolumn{5}{@{}c@{}}{\includegraphics[width=.9\linewidth]{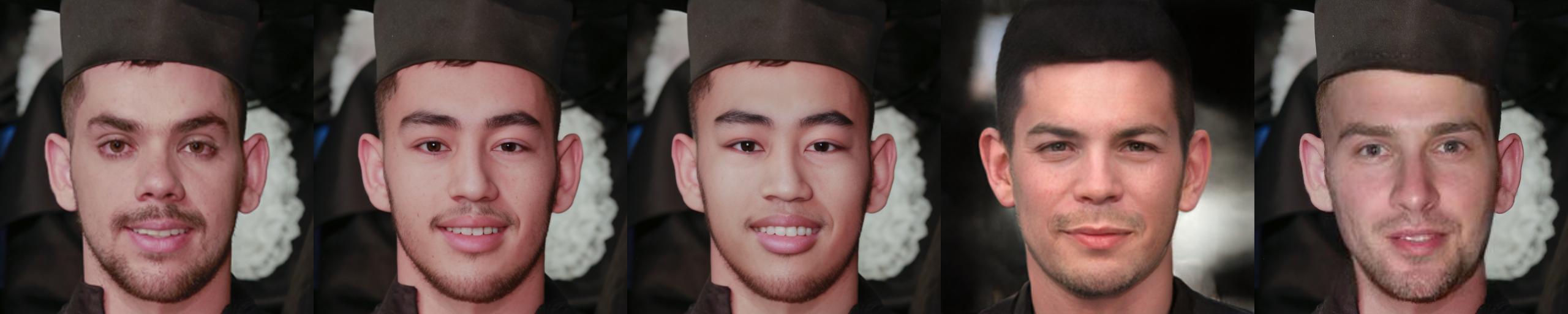}} \\
    \multicolumn{5}{@{}c@{}}{\includegraphics[width=.9\linewidth]{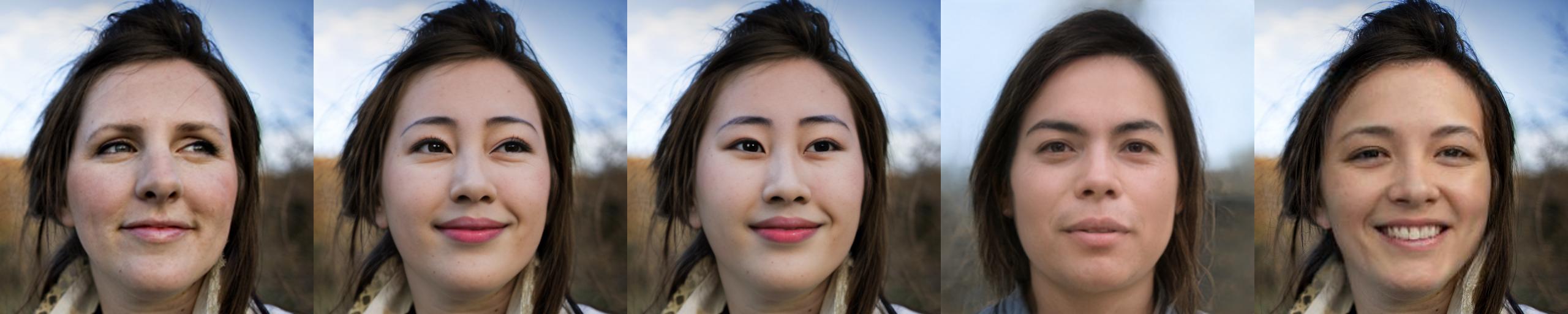}} \\
    \multicolumn{5}{@{}c@{}}{\includegraphics[width=.9\linewidth]{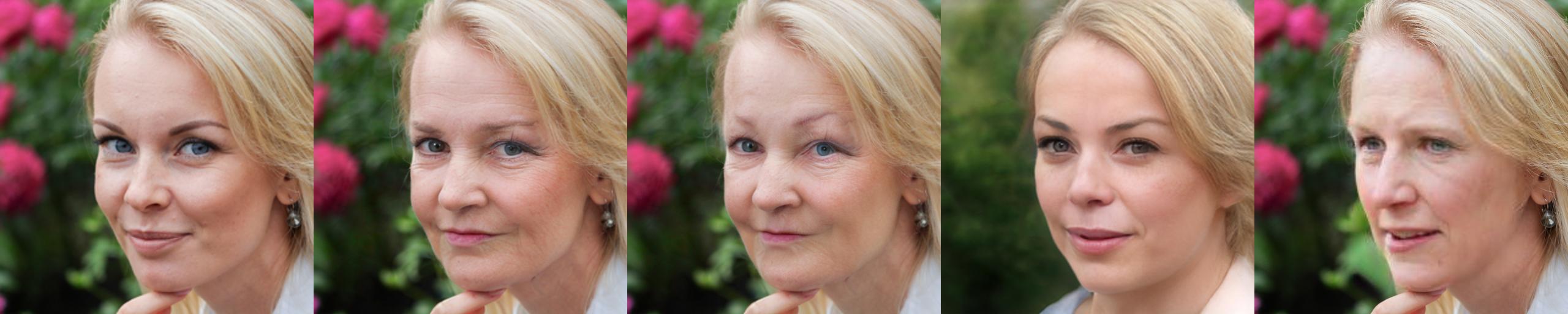}} \\
    \multicolumn{5}{@{}c@{}}{\includegraphics[width=.9\linewidth]{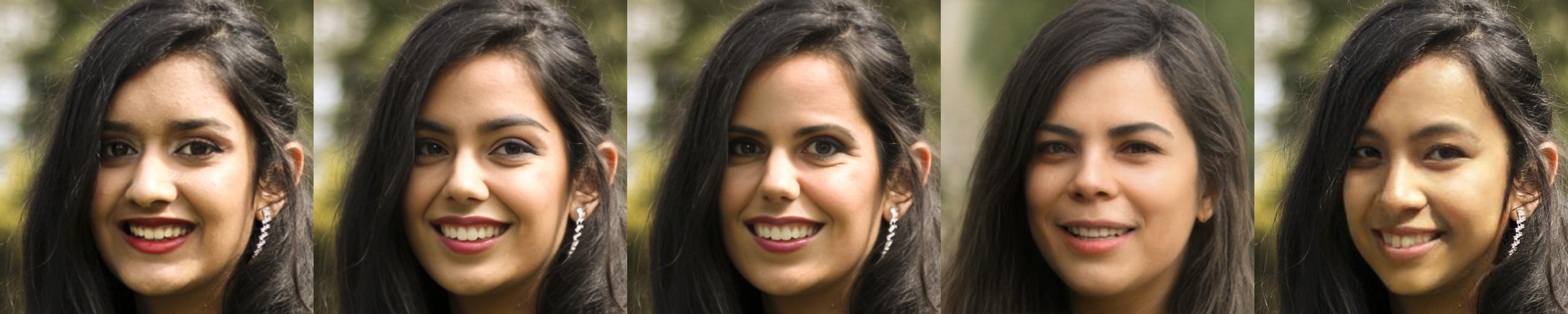}} \\
    \multicolumn{5}{@{}c@{}}{\includegraphics[width=.9\linewidth]{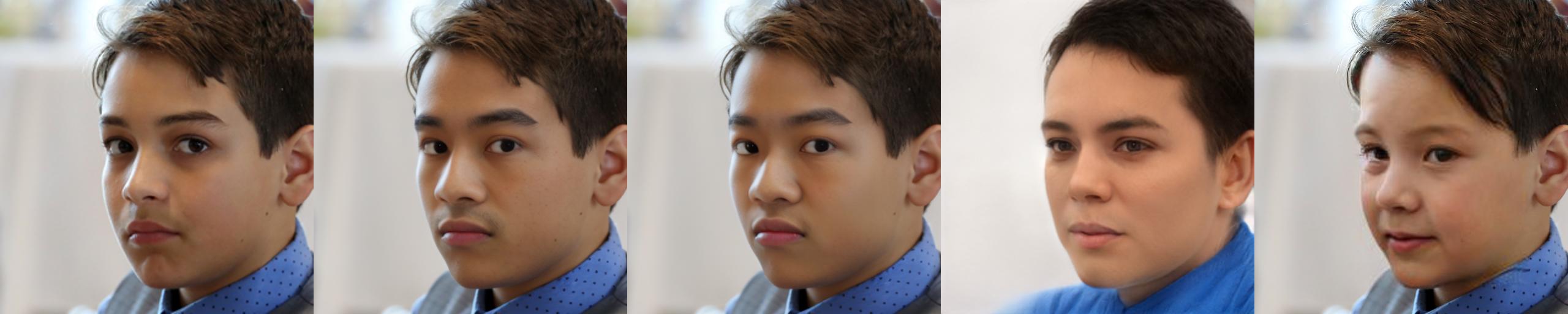}} \\
    Original & Ours \((d=1.2)\) & Ours \((d=1.4)\) & RiDDLE~\cite{li2023riddle} & DP2~\cite{hukkelaas2023deepprivacy2} \\
  \end{tabular}
  \caption{Qualitative results on the task of face anonymization for FFHQ~\cite{karras2019style} test set.}
  \label{fig:anon_qual_ffhq3}
\end{figure*}

\Cref{fig:anon_qual_cele1,fig:anon_qual_cele2,fig:anon_qual_cele3,fig:anon_qual_ffhq1,fig:anon_qual_ffhq2,fig:anon_qual_ffhq3} present additional qualitative results of our anonymization technique. We showcase these results using images from our test sets in FFHQ~\cite{karras2019style} and CelebA-HQ~\cite{karras2017progressive} databases. We also compare our method's performance against the same set of anonymization techniques~\cite{barattin2023attribute,hukkelaas2023deepprivacy2,li2023riddle} discussed earlier in our paper. 

\section*{Face Swapping Results}

\begin{figure*}
  \footnotesize
  \centering
  \begin{tabular}{*{6}{>{\centering\arraybackslash}m{\dimexpr.16\linewidth-2\tabcolsep}}}
    \multicolumn{6}{@{}c@{}}{\includegraphics[width=.96\linewidth]{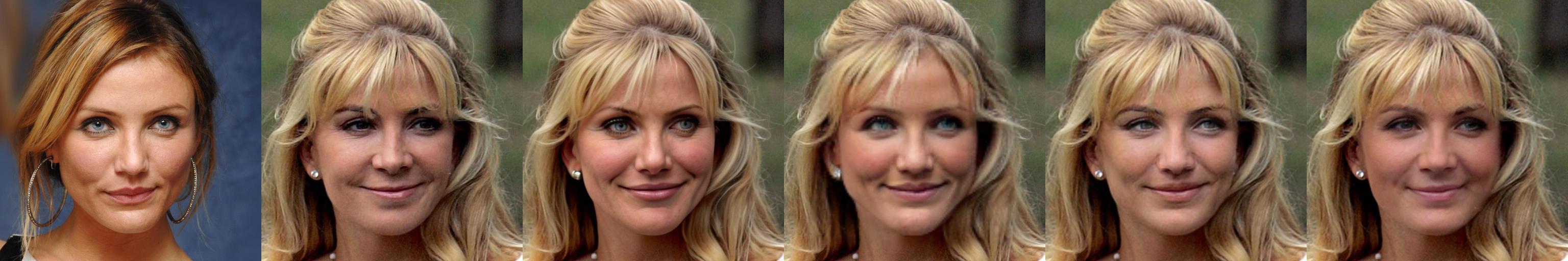}} \\
    \multicolumn{6}{@{}c@{}}{\includegraphics[width=.96\linewidth]{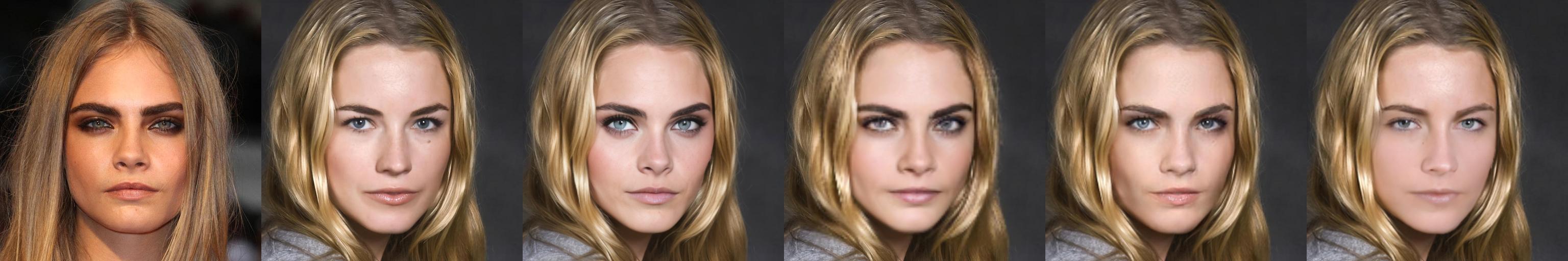}} \\
    \multicolumn{6}{@{}c@{}}{\includegraphics[width=.96\linewidth]{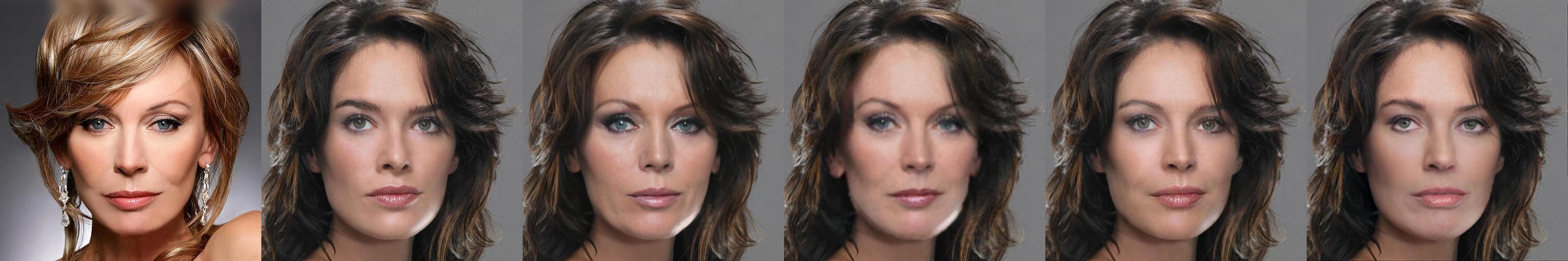}} \\
    \multicolumn{6}{@{}c@{}}{\includegraphics[width=.96\linewidth]{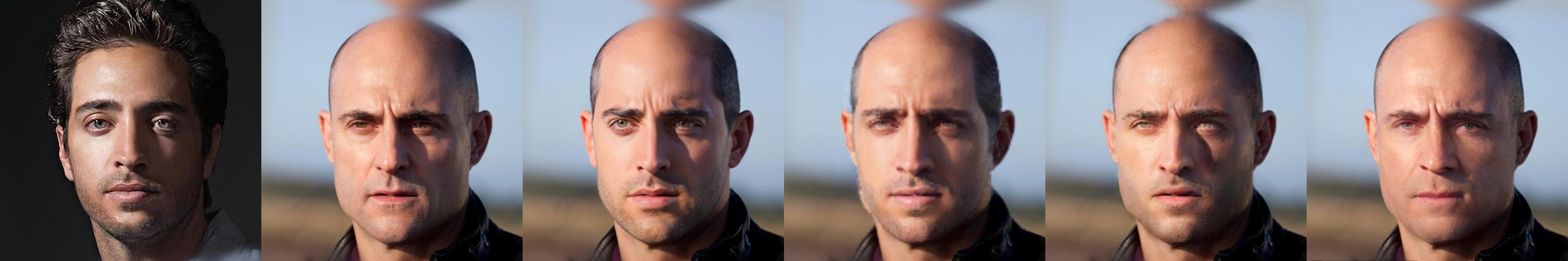}} \\
    \multicolumn{6}{@{}c@{}}{\includegraphics[width=.96\linewidth]{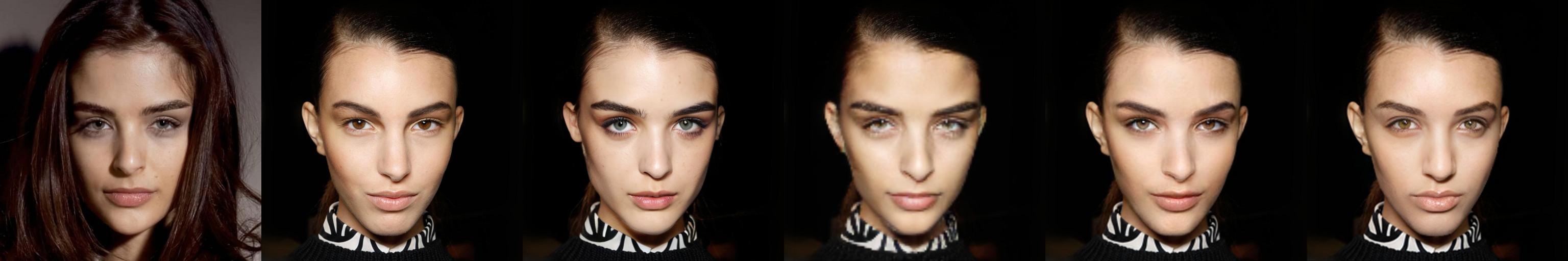}} \\
    \multicolumn{6}{@{}c@{}}{\includegraphics[width=.96\linewidth]{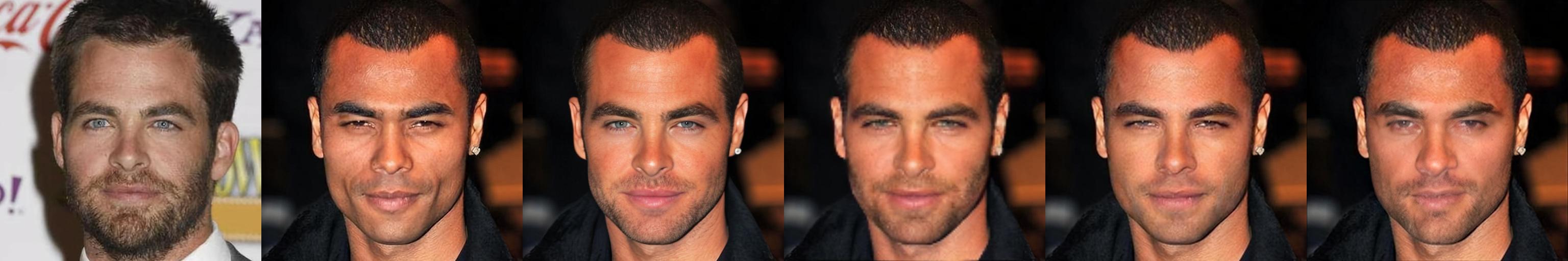}} \\
    Source & Driving & Ours & InSwapper~\cite{Guo_InsightFace_Swapper} & BlendFace~\cite{shiohara2023blendface} & DiffSwap~\cite{zhao2023diffswap} \\
  \end{tabular}
  \caption{Qualitative results on the task of face swapping for CelebA-HQ~\cite{karras2017progressive} test set.}
  \label{fig:swap_qual_cele1}
\end{figure*}

\begin{figure*}
  \footnotesize
  \centering
  \begin{tabular}{*{6}{>{\centering\arraybackslash}m{\dimexpr.16\linewidth-2\tabcolsep}}}
    \multicolumn{6}{@{}c@{}}{\includegraphics[width=.96\linewidth]{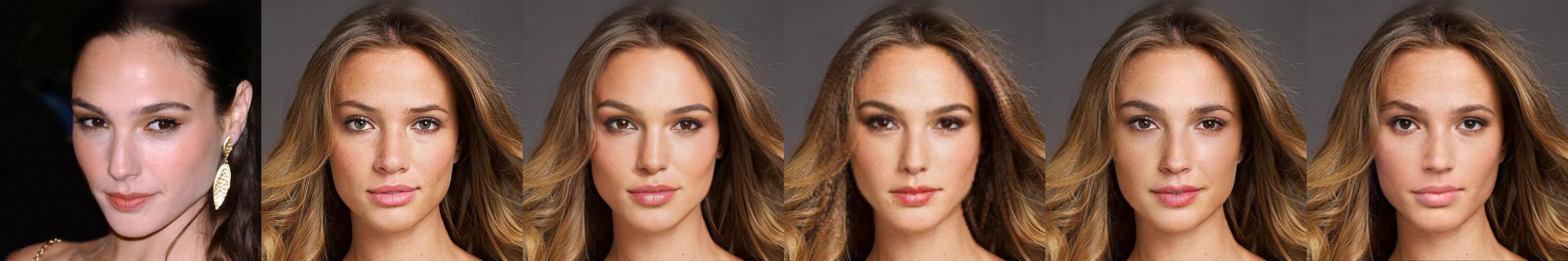}} \\
    \multicolumn{6}{@{}c@{}}{\includegraphics[width=.96\linewidth]{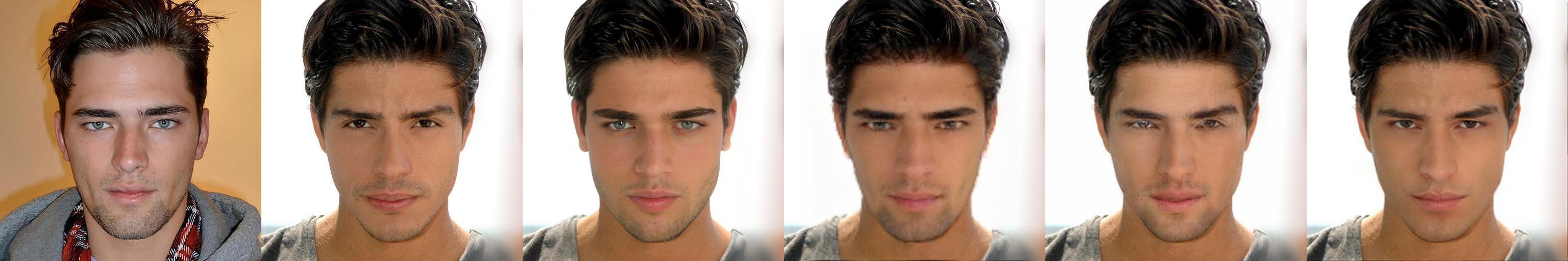}} \\
    \multicolumn{6}{@{}c@{}}{\includegraphics[width=.96\linewidth]{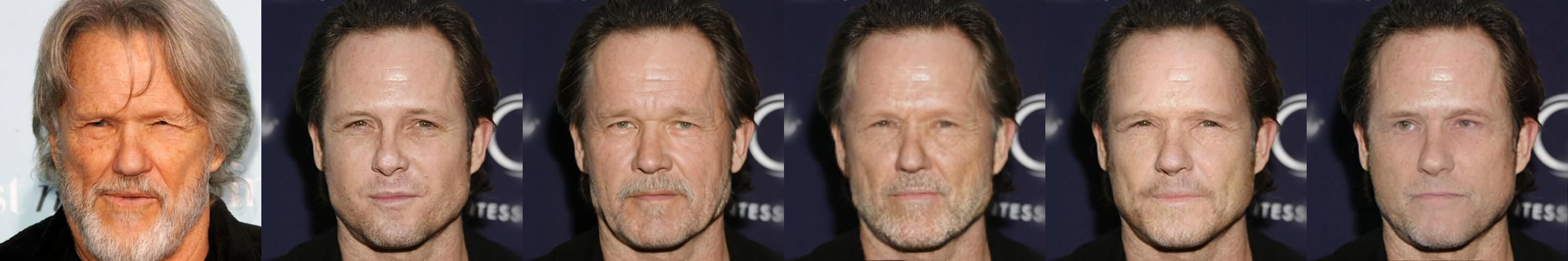}} \\
    \multicolumn{6}{@{}c@{}}{\includegraphics[width=.96\linewidth]{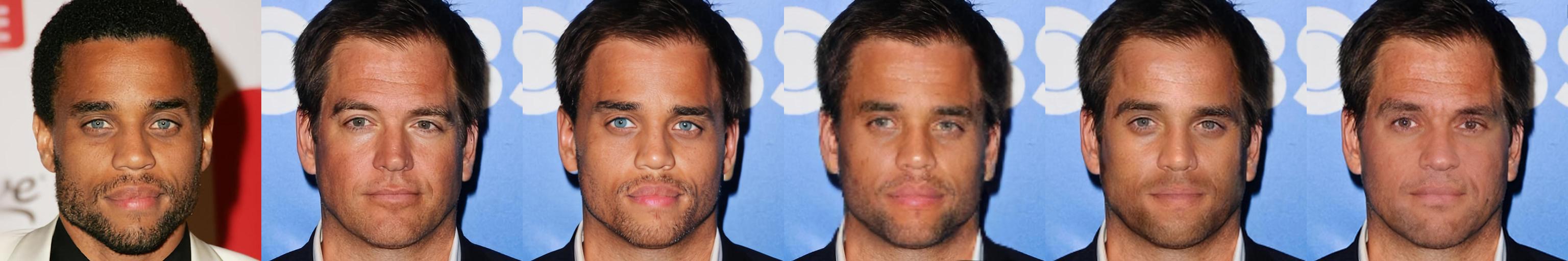}} \\
    \multicolumn{6}{@{}c@{}}{\includegraphics[width=.96\linewidth]{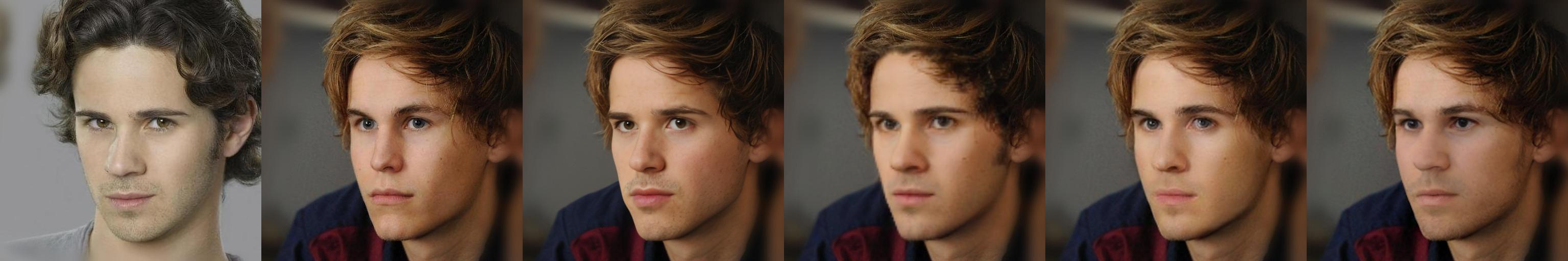}} \\
    \multicolumn{6}{@{}c@{}}{\includegraphics[width=.96\linewidth]{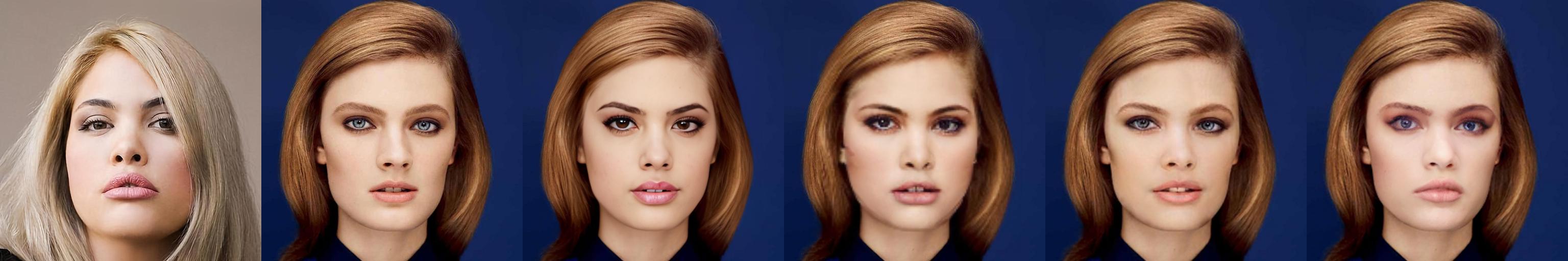}} \\
    Source & Driving & Ours & InSwapper~\cite{Guo_InsightFace_Swapper} & BlendFace~\cite{shiohara2023blendface} & DiffSwap~\cite{zhao2023diffswap} \\
  \end{tabular}
  \caption{Qualitative results on the task of face swapping for CelebA-HQ~\cite{karras2017progressive} test set.}
  \label{fig:swap_qual_cele2}
\end{figure*}

\begin{figure*}
  \footnotesize
  \centering
  \begin{tabular}{*{6}{>{\centering\arraybackslash}m{\dimexpr.16\linewidth-2\tabcolsep}}}
    \multicolumn{6}{@{}c@{}}{\includegraphics[width=.96\linewidth]{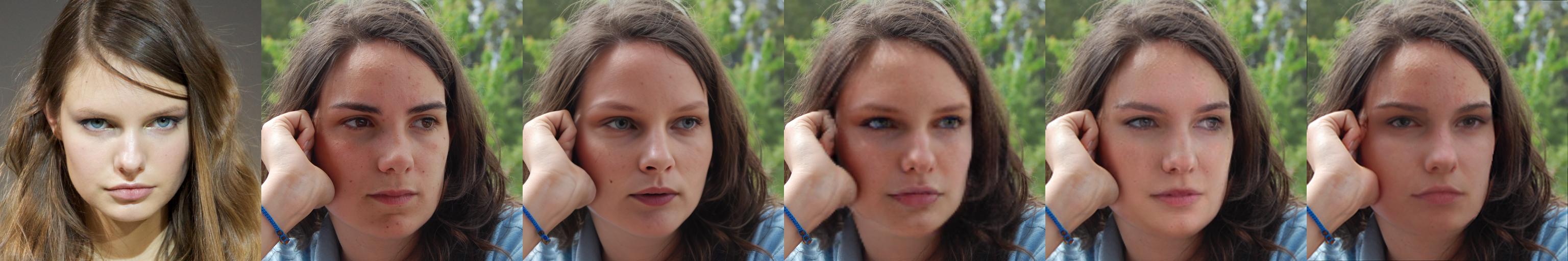}} \\
    \multicolumn{6}{@{}c@{}}{\includegraphics[width=.96\linewidth]{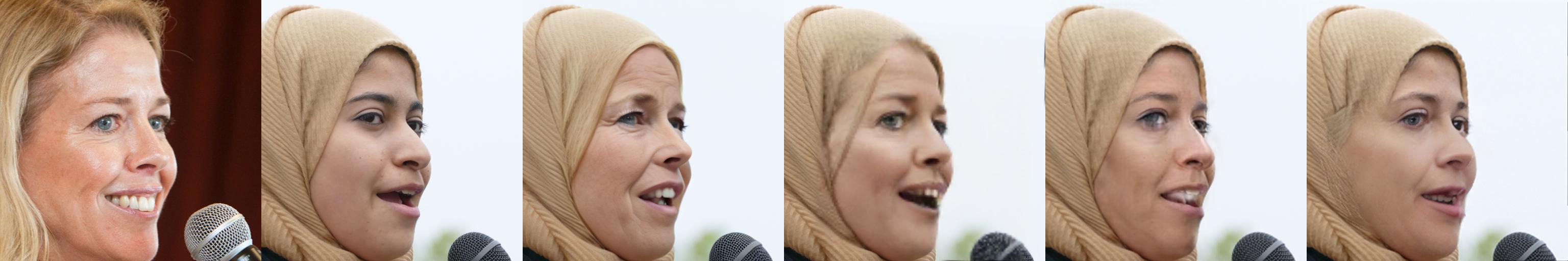}} \\
    \multicolumn{6}{@{}c@{}}{\includegraphics[width=.96\linewidth]{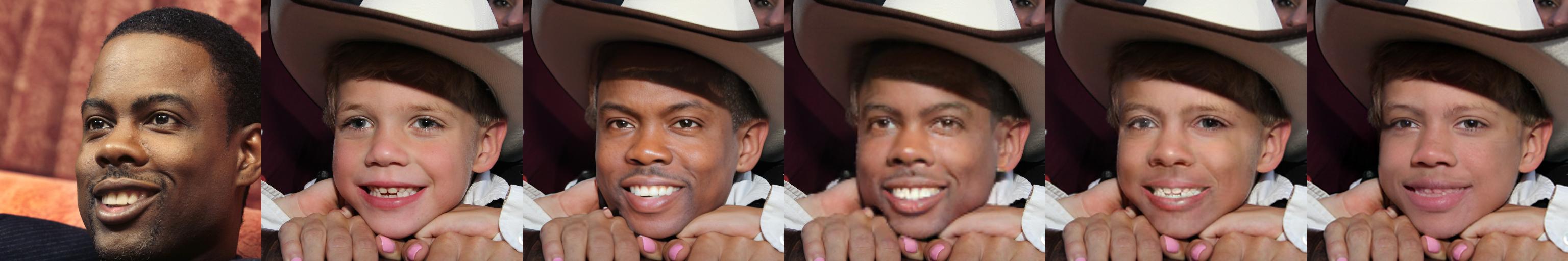}} \\
    \multicolumn{6}{@{}c@{}}{\includegraphics[width=.96\linewidth]{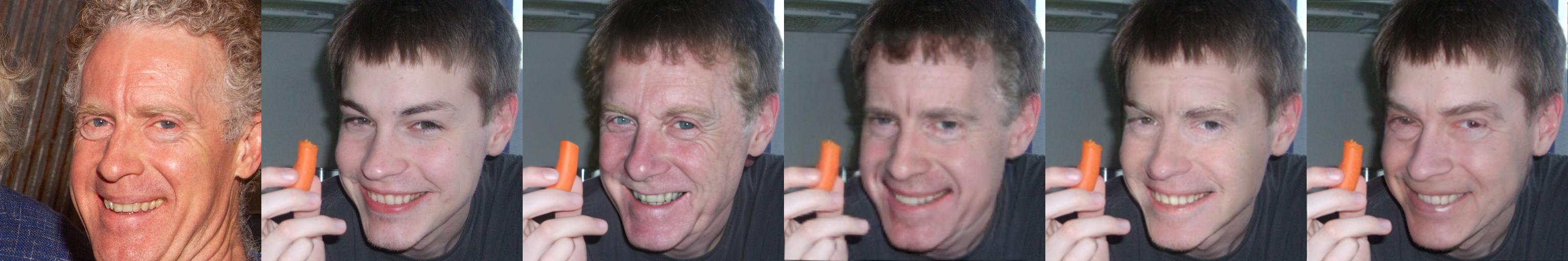}} \\
    \multicolumn{6}{@{}c@{}}{\includegraphics[width=.96\linewidth]{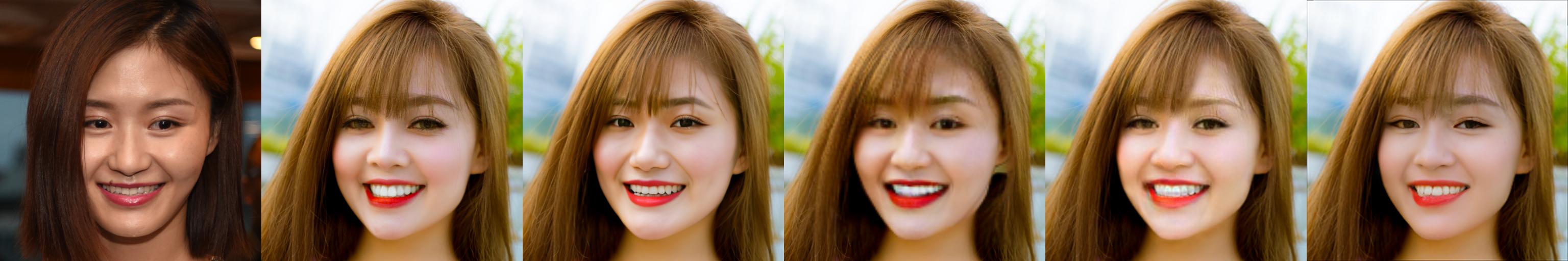}} \\
    \multicolumn{6}{@{}c@{}}{\includegraphics[width=.96\linewidth]{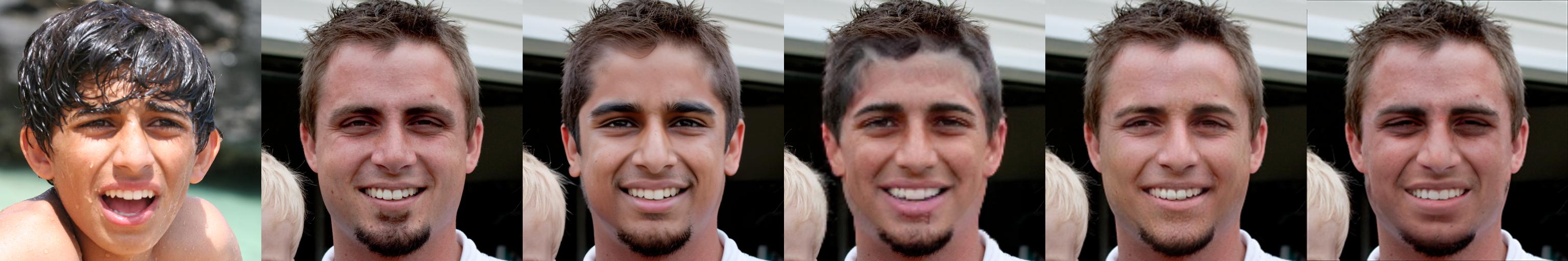}} \\
    Source & Driving & Ours & InSwapper~\cite{Guo_InsightFace_Swapper} & BlendFace~\cite{shiohara2023blendface} & DiffSwap~\cite{zhao2023diffswap} \\
  \end{tabular}
  \caption{Qualitative results on the task of face swapping for FFHQ~\cite{karras2019style} test set.}
  \label{fig:swap_qual_ffhq1}
\end{figure*}

\begin{figure*}
  \footnotesize
  \centering
  \begin{tabular}{*{6}{>{\centering\arraybackslash}m{\dimexpr.16\linewidth-2\tabcolsep}}}
    \multicolumn{6}{@{}c@{}}{\includegraphics[width=.96\linewidth]{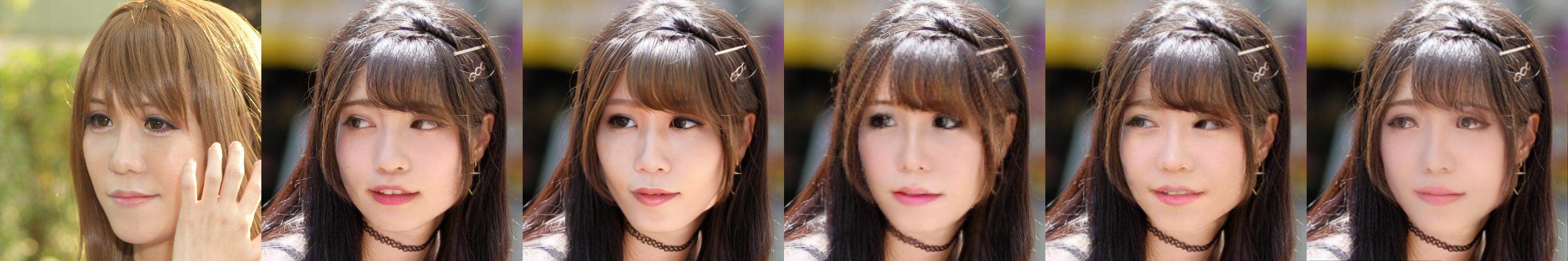}} \\
    \multicolumn{6}{@{}c@{}}{\includegraphics[width=.96\linewidth]{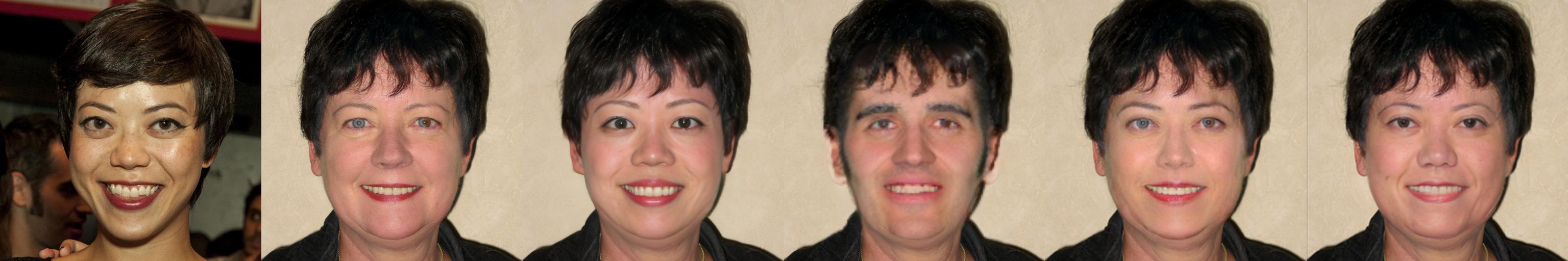}} \\
    \multicolumn{6}{@{}c@{}}{\includegraphics[width=.96\linewidth]{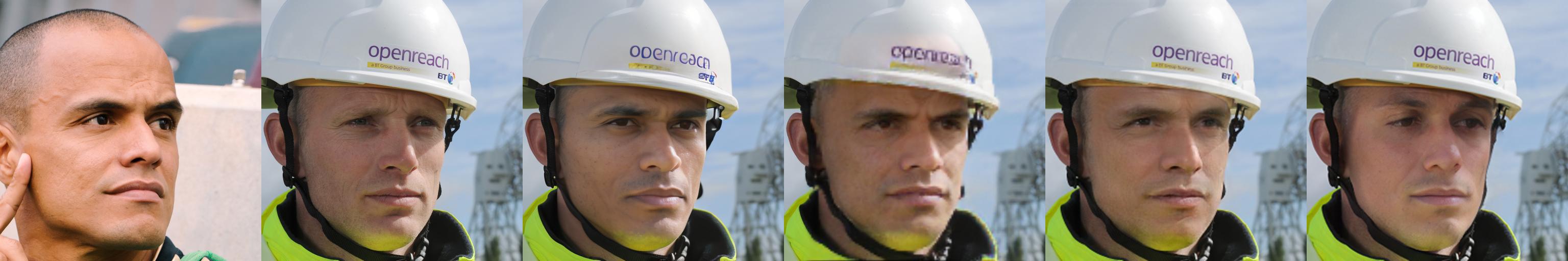}} \\
    \multicolumn{6}{@{}c@{}}{\includegraphics[width=.96\linewidth]{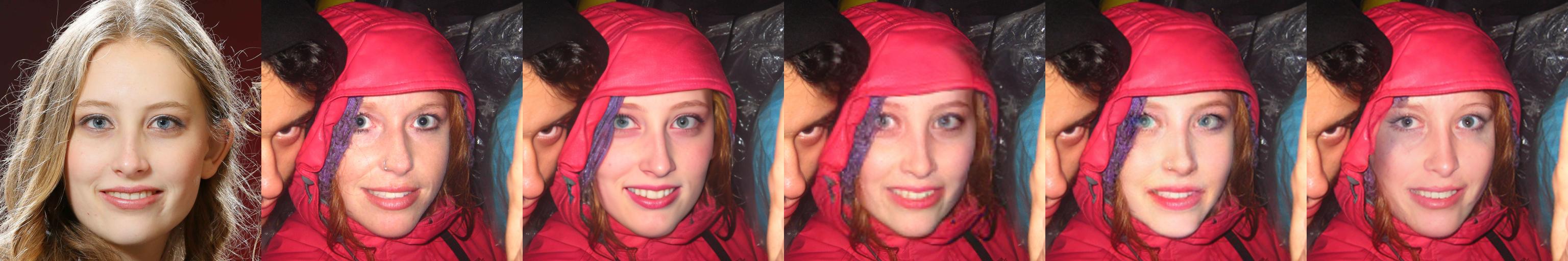}} \\
    \multicolumn{6}{@{}c@{}}{\includegraphics[width=.96\linewidth]{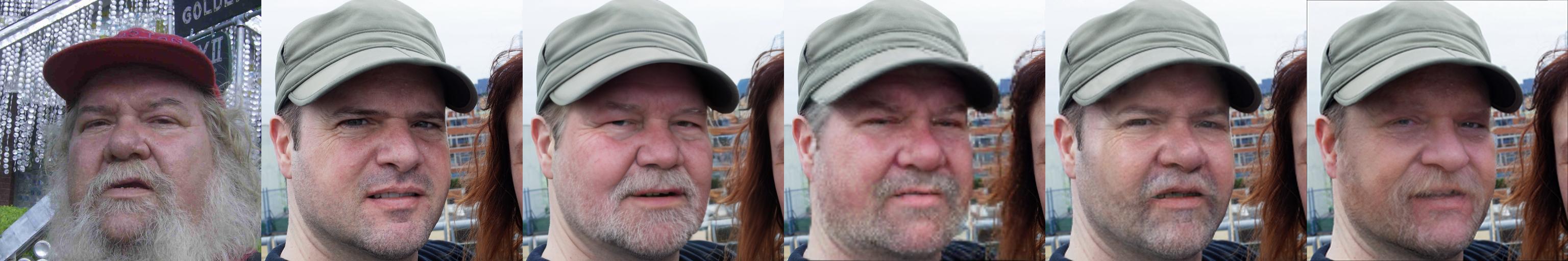}} \\
    \multicolumn{6}{@{}c@{}}{\includegraphics[width=.96\linewidth]{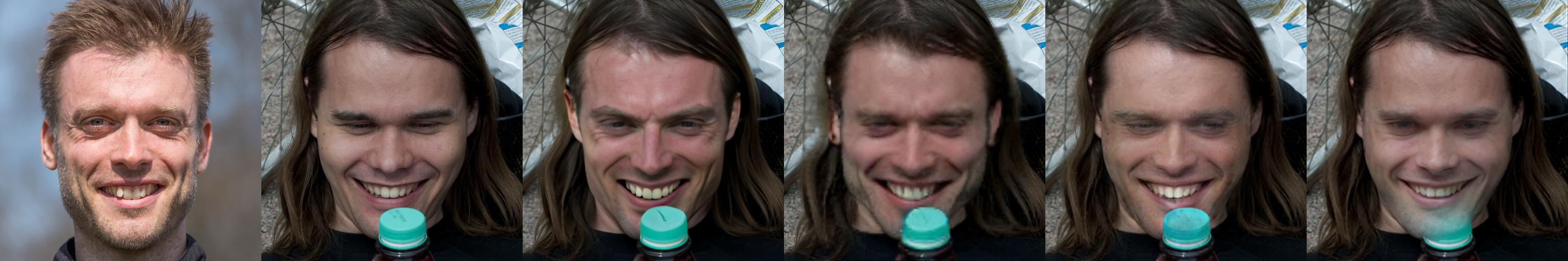}} \\
    Source & Driving & Ours & InSwapper~\cite{Guo_InsightFace_Swapper} & BlendFace~\cite{shiohara2023blendface} & DiffSwap~\cite{zhao2023diffswap} \\
  \end{tabular}
  \caption{Qualitative results on the task of face swapping for FFHQ~\cite{karras2019style} test set.}
  \label{fig:swap_qual_ffhq2}
\end{figure*}

Although face swapping is not the primary focus of our research, our model initially develops this capability as part of its anonymization process. To demonstrate its effectiveness, we present additional face swapping examples using the FFHQ~\cite{karras2019style} and CelebA-HQ~\cite{karras2017progressive} datasets in ~\cref{fig:swap_qual_cele1,fig:swap_qual_cele2,fig:swap_qual_ffhq1,fig:swap_qual_ffhq2}. We also compare our results to established face swapping benchmarks~\cite{zhao2023diffswap,Guo_InsightFace_Swapper,shiohara2023blendface} discussed in our paper. These comparisons showcase our model's superior ability to generate high-quality facial images.

Furthermore, \cref{tab:swap_quan} provides quantitative results for the face swapping tasks. These results indicate that our model achieves superior Image Quality Assessment (IQA) scores across both datasets. While both DiffSwap~\cite{zhao2023diffswap} and our model can natively generate high-resolution images at 512 $\times$ 512, our model achieves an IQA score that is more than 30\% higher than DiffSwap's~\cite{zhao2023diffswap}. This improvement is likely due to our use of ReferenceNet, which encodes fine-grained features and enables our model to produce higher quality facial images.

\section*{Societal Impact of AI-Generated Faces}

AI-generated faces present a dual challenge in our digital world. While they can enhance privacy by offering anonymity, they also create opportunities for malicious activities. Scammers might use these synthetic identities to produce more convincing deceptions, potentially eroding trust in online interactions and media. To address these risks, a comprehensive strategy is essential. This includes technological solutions such as advanced watermarking and AI detection systems, along with legal frameworks regulating the use of synthetic faces. Additionally, raising public awareness about this technology and its potential misuse is crucial. Establishing clear industry standards for the ethical creation and application of AI-generated faces will help balance their benefits while protecting social trust. A coordinated effort across technological, legal, and educational fronts is vital for maximizing the positive potential of this innovation while minimizing its societal drawbacks.

\end{document}